\documentclass[final]{cvpr}
\usepackage{bm}
\usepackage{times}
\usepackage{graphicx}
\usepackage{graphbox}
\usepackage{amsmath}
\usepackage{amssymb}
\usepackage{caption}
\usepackage{subfigure}

\makeatletter 
\renewcommand{\@thesubfigure}{\hskip\subfiglabelskip}
\makeatother

\usepackage[pagebackref=true,breaklinks=true,colorlinks,bookmarks=false]{hyperref}

\def\cvprPaperID{xxx} 
\def\confYear{CVPR 2021}

\begin{document}

\title{Learning Heatmap-Style Jigsaw Puzzles Provides Good\\Pretraining for 2D Human Pose Estimation}

\author{Kun Zhang$^{1,2}$\footnotemark[1] , Rui Wu$^{3}$\thanks{Equal Contributions.} , Ping Yao$^{1,2}$\thanks{Corresponding Author.} , Kai Deng$^{1,2}$, Ding Li$^{2,4}$, Renbiao Liu$^{5}$,\\ Chuanguang Yang$^{1,2}$, Ge Chen$^{1,2}$, Min Du$^{3}$, and Tianyao Zheng$^{1,2}$\\
$^1$Institute of Computing Technology, Chinese Academy of Sciences\\
$^2$University of Chinese Academy of Sciences
$^3$Horizon Robotics\\
$^4$Institute of Automation, Chinese Academy of Sciences
$^5$Harbin Institute of Technology\\
{\tt\small zhangkun@ieee.org \{yaoping,dengkai19s,yangchuanguang,chenge18s,mailtozty\}@ict.ac.cn}\\
{\tt\small \{rui.wu,min.du\}@horizon.ai \tt\small liding2016@ia.ac.cn richardodliu@gmail.com}

}

\maketitle

\begin{abstract}
The target of 2D human pose estimation is to locate the keypoints of body parts from input 2D images. State-of-the-art methods for pose estimation usually construct pixel-wise heatmaps from keypoints as labels for learning convolution neural networks, which are usually initialized randomly or using classification models on ImageNet as their backbones. We note that 2D pose estimation task is highly dependent on the contextual relationship between image patches, thus we introduce a self-supervised method for pretraining 2D pose estimation networks. Specifically, we propose Heatmap-Style Jigsaw Puzzles (HSJP) problem as our pretext-task, whose target is to learn the location of each patch from an image composed of shuffled patches. During our pretraining process, we only use images of person instances in MS-COCO, rather than introducing extra and much larger ImageNet dataset. A heatmap-style label for patch location is designed and our learning process is in a non-contrastive way. The weights learned by HSJP pretext task are utilised as backbones of 2D human pose estimator, which are then finetuned on MS-COCO human keypoints dataset. With two popular and strong 2D human pose estimators, HRNet and SimpleBaseline, we evaluate mAP score on both MS-COCO validation and test-dev datasets. Our experiments show that downstream pose estimators with our self-supervised pretraining obtain much better performance than those trained from scratch, and are comparable to those using ImageNet classification models as their initial backbones.
\let\thefootnote\relax\footnote{Preprint Under Review.}
\end{abstract}

\section{Introduction}
2D human pose estimation refers to the task of locating important human body parts (e.g. elbow, head, ankle, wrist, shoulder) from input images, which serves as a basic task for varieties of computer vision applications, such as person re-id \cite{reid}, video analysis \cite{vid}, action recognition \cite{stgcn}, and human-computer interaction \cite{posetrack}. 

In recent years, the progress of 2D pose estimation has been greatly benefited by CNNs \cite{resnet,deeppose,ai20}. There are three mainstreams for this task: (1) top-down methods \cite{cpn1,hrv1} firstly separate all person instances with person detector, and then localise joints for each person instance with single-person pose estimation model; (2) bottom-up approaches \cite{openpose,pifpaf} firstly utilise joint estimator to localise all person joints, and then assign them to different person instances; (3) single-stage methods \cite{spm,directpose} predict person instances and human keypoints simultaneously. These three types of methods are suitable for different circumstances, but state-of-the-art performances usually come from heatmap-based top-down approaches \cite{lecunpose}, such as HRNet-based Networks \cite{udppose,hrv1,darkpose}, Pyramid-based Networks \cite{rsn,cpn1,mspn,seu2019}, and Hourglass Networks \cite{hourglasse1,hourglass}. These methods utilise CNNs to predict pixel-wise 2D heatmap for accurate human keypoint localisation. Our paper is focused on the pretraining of single-person pose estimator in the top-down framework.


\begin{figure*}
\centering
\subfigure[(a) Initial Image Patches]{
\includegraphics[width=.287\textwidth]{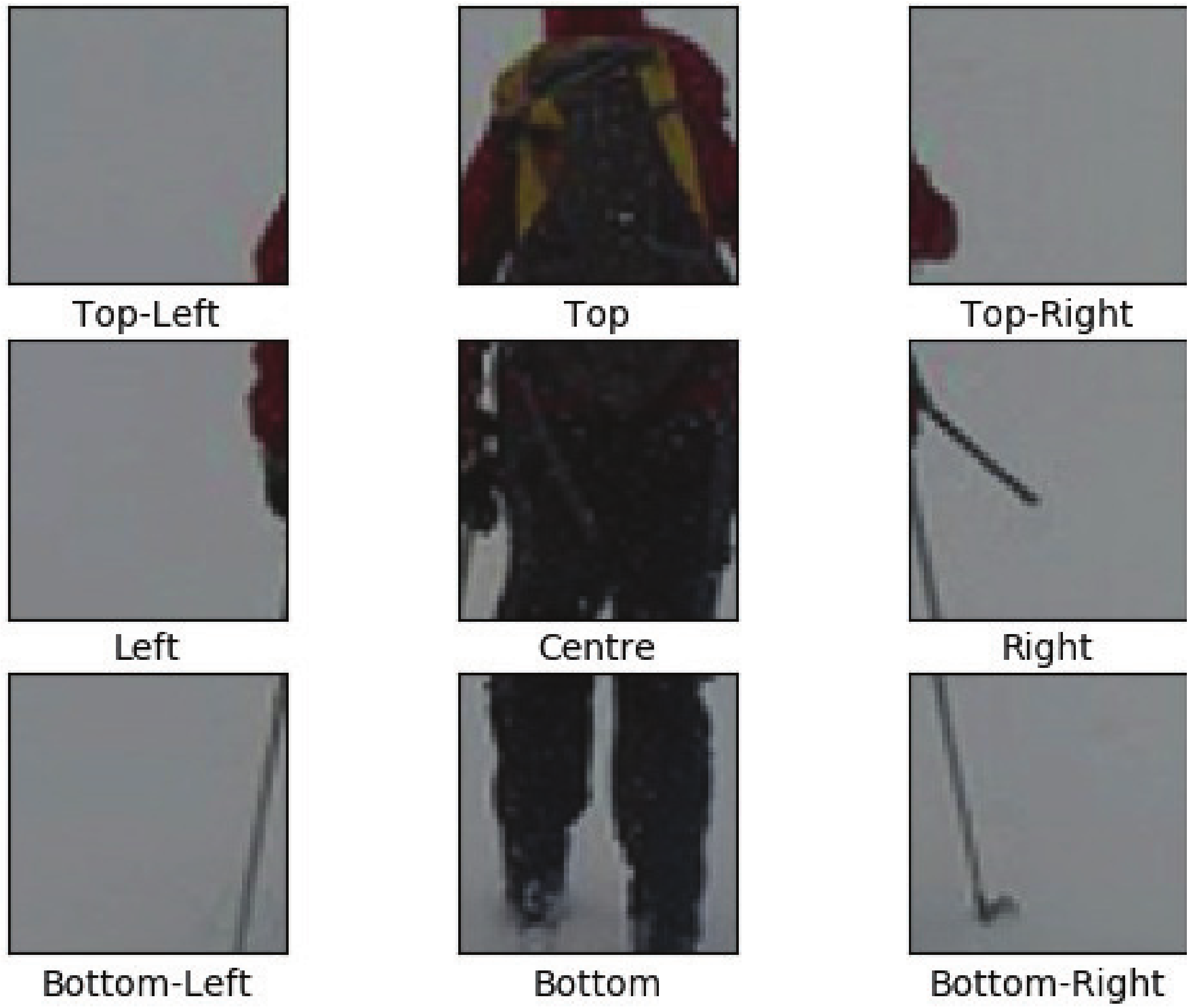}
\label{initialimage}
}
\hspace{.042\textwidth}
\subfigure[(b) Shuffled Image Patches]{
\includegraphics[width=.287\textwidth]{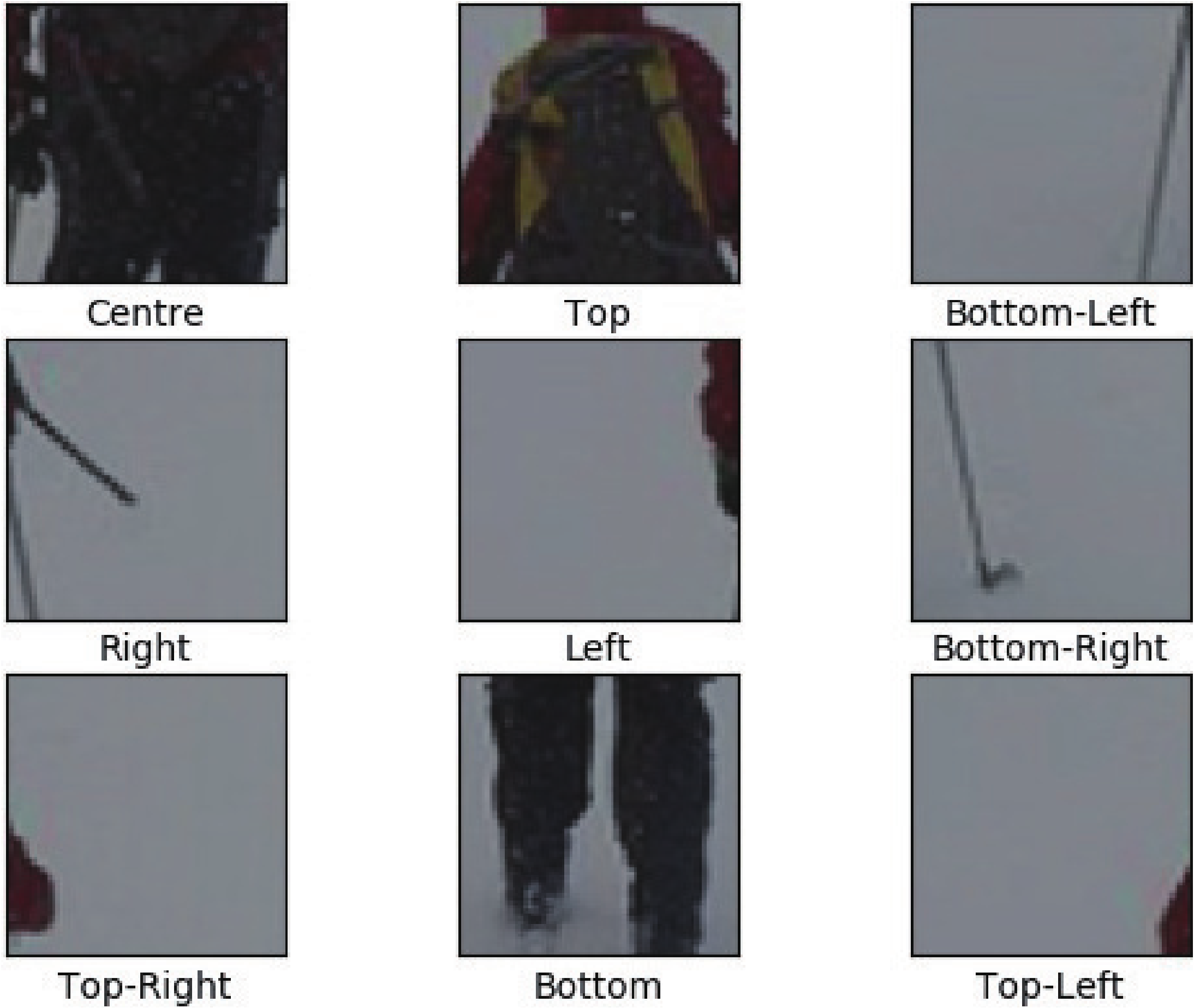}
\label{shuffledimage}
}
\hspace{.042\textwidth}
\subfigure[(c)Target Heatmaps]{
\includegraphics[width=.287\textwidth]{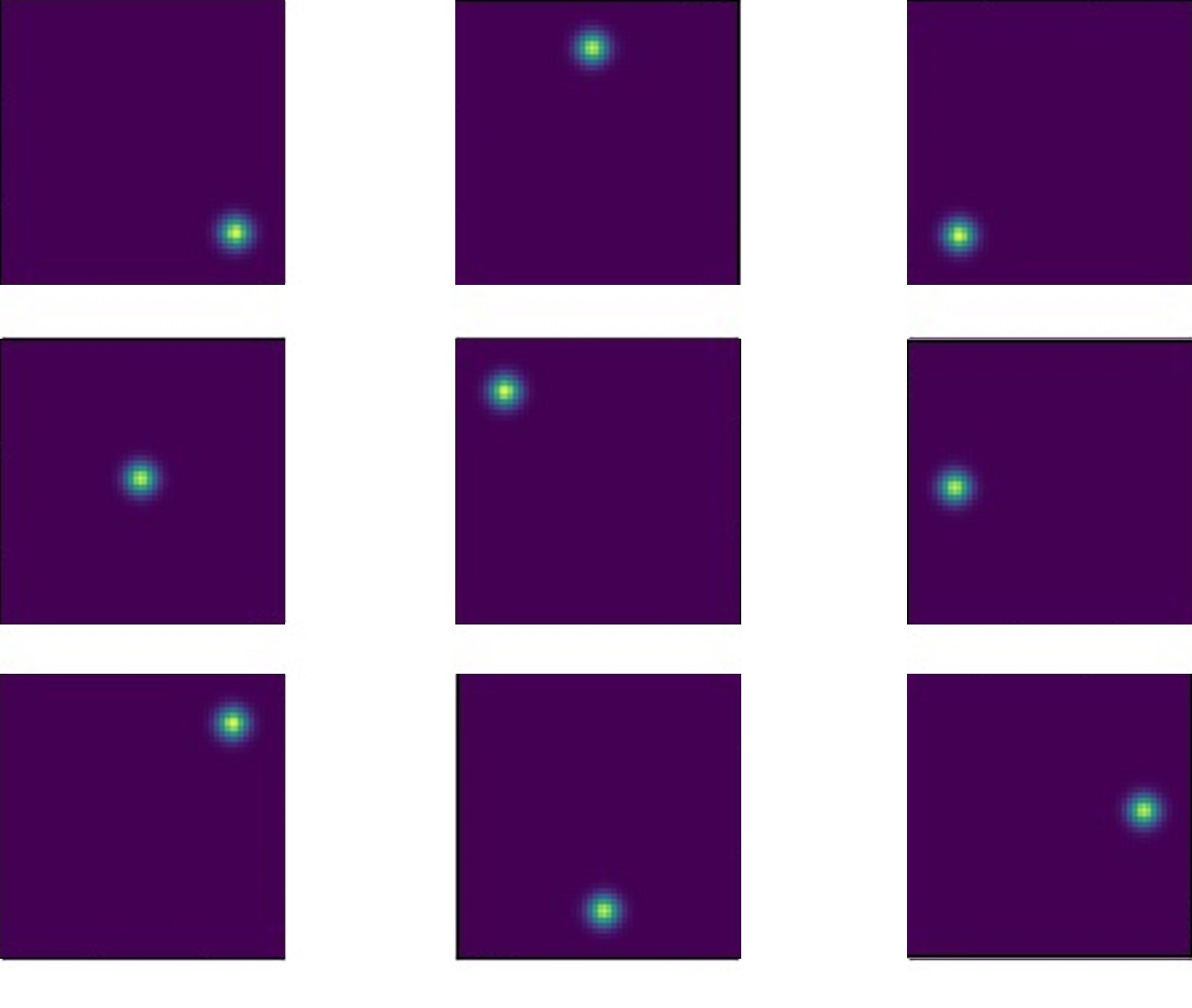}
\label{groundtruthmap}
}
\caption{Description of our proposed Heatmap-Style Jigsaw Puzzles task. (a) Splitting the original image into $N\times N$ patches, $N=3$ in this figure for clear demonstration. (b) Randomly shuffling the split $N\times N$ patches to build jigsaw puzzles problem. (c) Ground truth heatmaps for channels corresponding to their image patches.} 
\label{taskdescript}
\end{figure*}

While self-supervised learning has recently achieved significant progress in varieties of computer vision tasks \cite{miccaiself,simclr,mocov1,dengjiaself,vipl20seg}, supervised methods are still dominant for the pretraining of 2D pose estimation. Instead of finetuning from ImageNet classification weights \cite{imgnet,hrv1,simplebaseline}, we design Heatmap-Style Jigsaw Puzzles (HSJP) problem as an efficient self-supervised learning pretext task for 2D human pose estimation. The description of our HSJP task is demonstrated in Figure \ref{taskdescript}. During the preprocessing of our HSJP task, we split the image of each person instance into $N\times N$ patches and shuffle them, which are illustrated in Figure \ref{initialimage} and \ref{shuffledimage}. As we have mentioned that the target of 2D pose estimation is to localise positions of human keypoints from input images, and in order to make our pretext task as similar to its downstream task as possible, the goal of HSJP task is to estimate the shuffled place of each original image patch with a heatmap-based approach. As Figure \ref{groundtruthmap} shows, the ground truth of our heatmap for the channel corresponding to each original image patch is a 2D Gaussain distribution located in the shuffled centre. For example, the top-left patch in Figure \ref{initialimage} is located in the bottom-right of Figure \ref{shuffledimage}; therefore, the peak of Gaussian Distribution in the first channel is located in its bottom-right. In this way, the target of our HSJP pretext task is quite similar to those of many state-of-the-art 2D pose estimation methods: all of them predict target locations by heatmap-based means. 

During the pretraining process of our HSJP task, the network learns to predict Gaussian peaks as centres of those shuffled patches for correspondences to solve the jigsaw puzzles problem: the image patch on the Gaussian peak of the $i$th channel in Figure \ref{groundtruthmap} should be put back to the $i$th patch in Figure \ref{initialimage}, $i\in [0, N^2)$. These correspondences learnt in HSJP task are also helpful to the downstream task of 2D human pose estimation, for the case that knowing relative positions of image patches is also beneficial to the localisation of human keypoints. For instance, if we know whether each shuffled patch should be the top-left or top-right corner in Figure \ref{taskdescript}, it's also helpful to locate shoulders of given person instance, as person shoulders are frequently located in neighbourhoods of top-left or top-right corners in original images before data augmentation.

Instead of supervised pretraining on labelled ImageNet dataset, our network is only pretrained with self-supervised HSJP pretext task on all person instances of MS-COCO \cite{coco} dataset, and finetuned on their human keypoint labels. Although we make no use of extra labels for pretraining, our keypoint performance is comparable to that of ImageNet pretraining on MS-COCO validation and test-dev datasets.

\section{Related Works}
\subsection{Heatmap-based human pose estimation}
Since \cite{lecunpose} firstly proposed to utilise a heatmap-based method for single-person pose estimation, recent state-of-the-art performances have mostly been achieved by relative means. For example, Hourglass-based Networks  \cite{hourglasse1,hourglass} apply symmetric network structures with skip connections to generate pixel-wise feature maps for accurate keypoint prediction; Pyramid-based Networks \cite{cpn1,seu2019} consolidate feature maps across ResNet stages, and some improvements were achieved by repeating pyramid architecture \cite{rsn,mspn}; HRNet \cite{hrv1} keeps high-resolution representations and exchanges information across multi-resolution subnetworks, and some recent works \cite{udppose,darkpose} focus on improving the preprocessing and postprocessing details of their codebase.

\subsection{Self-supervised pretraining for computer vision}
Self-supervised pretraining methods aim at designing pretext tasks without using handcrafted annotations so as to benefit their downstream targets. The main idea of self-supervised representation learning is to utilise annotations that are freely available from original datasets. It also saves labour costs of labelling datasets for machine learning. The past few years have witnessed the springing up of self-supervised means for computer vision field. A variety of pretext tasks have been proposed, such as spatial position prediction \cite{jigsaw19,gupta15iccv,jigsaw16}, transformation prediction \cite{tpred,vipl20seg}, image inpainting \cite{contextencoder}, image colourisation \cite{autocolor,colortrack,colorful16}, and learning latent embeddings \cite{simclr,boyl,mocov1,videoself,embed19}. Recently, some papers also discussed the robustness and usefulness of self-supervised pretraining approaches \cite{selfrobust,dengjiaself}. 

Applications on various computer vision tasks have also been discovered by self-supervised learning researchers in recent years. For examples, \cite{miccaiself} predicts boxes constructed from dataset to benefit cardiac MR image analysis, and \cite{vipl20seg} utilises self-supervised attention for weakly supervised semantic segmentation. However, supervised pretraining methods on large datasets such as ImageNet \cite{imgnet,hrv1,simplebaseline} are still dominant for 2D human pose estimation. Starting from the status quo, we design to solve Heatmap-Style Jigsaw Puzzles (HSJP) as pretext task for pretraining heatmap-based 2D pose estimation networks, and we have achieved competitive performance on MS-COCO benchmark.

\subsection{Methods for solving jigsaw puzzles}
The target of jigsaw puzzles problem is to put shuffled image patches back to their original places, which serves as an effective pretext for learning visuospatial dependencies in real world \cite{realjigsaw}. Some previous works \cite{jigsaw19,jigsaw16} also made use of jigsaw puzzles to learn vector-based representations for computer vision tasks such as image classification or object detection, but never had we encountered any paper that solve jigsaw puzzles as the pretext task for 2D human pose estimation methods: they are usually non-pretrained \cite{rsn,mspn,hourglass}, or pretrained on ImageNet dataset \cite{imgnet,hrv1,simplebaseline}.

A large number of papers focus on providing accurate methods for solving jigsaw puzzles. Early works \cite{apictorial,jigsaw94pr} consider using information from shape or appearance to solve this problem. Recently, there are two mainstreams for reassembling image patches: greedy methods \cite{jigsawcv12,jigsawcv18,jigsawcv14} and global methods \cite{jigsawcvpr20,jigsawcvpr10,jigsawcvpr16}. We don't concentrate on proposing any accurate method for solving jigsaw puzzles in this paper, but aim at learning suitable representations for pretraining heatmap-based 2D pose estimation networks. 

\section{Heatmap-Style Jigsaw Puzzles}
As is mentioned in the introduction, the target of 2D pose estimation is to localise positions of human keypoints from 2D input images. In order to make our self-supervised pretraining target as relative to downstream task as possible, we propose Heatmap-Style Jigsaw Puzzles (HSJP) as pretext task for self-supervised pretraining, whose goal is to estimate the shuffled locations of each original image patch. The pretext and downstream tasks both predict certain part of a person instance (patch or keypoint) with a heatmap-based approach. In this section, we will elaborate details of how to pretrain a 2D human pose estimator with HSJP task. 

\subsection{Target description for HSJP pretext task}
The target of our proposed HSJP pretext is to estimate the shuffled place of each original image patch with a heatmap-based approach, whose inputs include all person instances cropped from COCO dataset. To make our self-supervised pretraining method comparable to ImageNet \cite{imgnet,hrv1,simplebaseline} pretraining, the resolution of input is $224\times 224$, which is the same as most popular classification methods on ImageNet dataset. We separate input image $I$ into $N\times N$ patches of $\lceil\frac{224}{N}\rceil\times\lceil\frac{224}{N}\rceil$ during data preprocessing, as is illustrated in Figure \ref{initialimage}. We denote the index of each split image patch as $i\in [0, N^2)$, which is also corresponding to the channel of output heatmap. The size of each image patch is rounded up for the case where $N$ cannot divide 224. 

We then apply Knuth-Durstenfeld algorithm \cite{dshuffle,knuth1997} to randomly shuffle the split image patches, as is illustrated in Figure \ref{shuffledimage}. The centre of each shuffled image patch is denoted as $\bm p_i=(x_i,y_i)$. In order to make the target $\bm G$ of pretext task as similar to 2D human pose estimation as possible, the ground truth heatmap for each output channel $i$ is a 2-dimensional Gaussian Distribution located on each patch centre $\bm p_i$. For each output channel $i$, the heatmap response at location $\bm p'=(x', y')$ is computed as:
\begin{equation}
\bm G_{x'y'}^i=\exp(-\frac{(x'-x_i)^2+(y'-y_i)^2}{2\sigma^2}),
\end{equation}
where $\sigma$ represents the standard deviation of 2D Gaussian Distribution, and $\sigma$ should be less than $\frac{224}{6N}$ according to $3\sigma$ rule to avoid mislocation. 

The network learns to predict 2D Gaussian heatmaps to obtain positions of shuffled image patches during the pretraining. We regress target heatmap by the same network as downstream task, except that the number of output channel is equal to $N\times N$: the number of image patches rather than the number of human keypoints. Our best $N$ for downstream performance is 6, which will be tested by ablation studies in Section \ref{patchesub}.
\label{tardis}

\subsection{Learning rich representations}
\begin{figure}
\centering
\subfigure[Image Patches]{
\includegraphics[width=.105\textwidth]{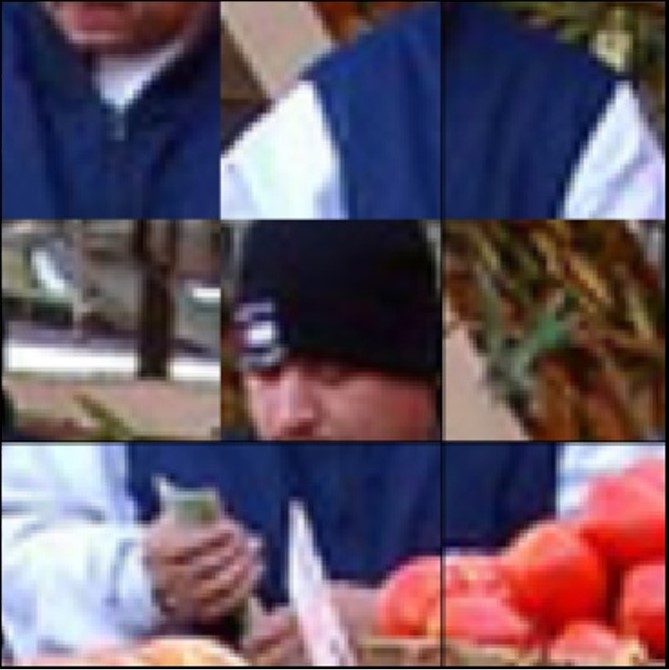}
}
\subfigure[Translated]{
\includegraphics[width=.105\textwidth]{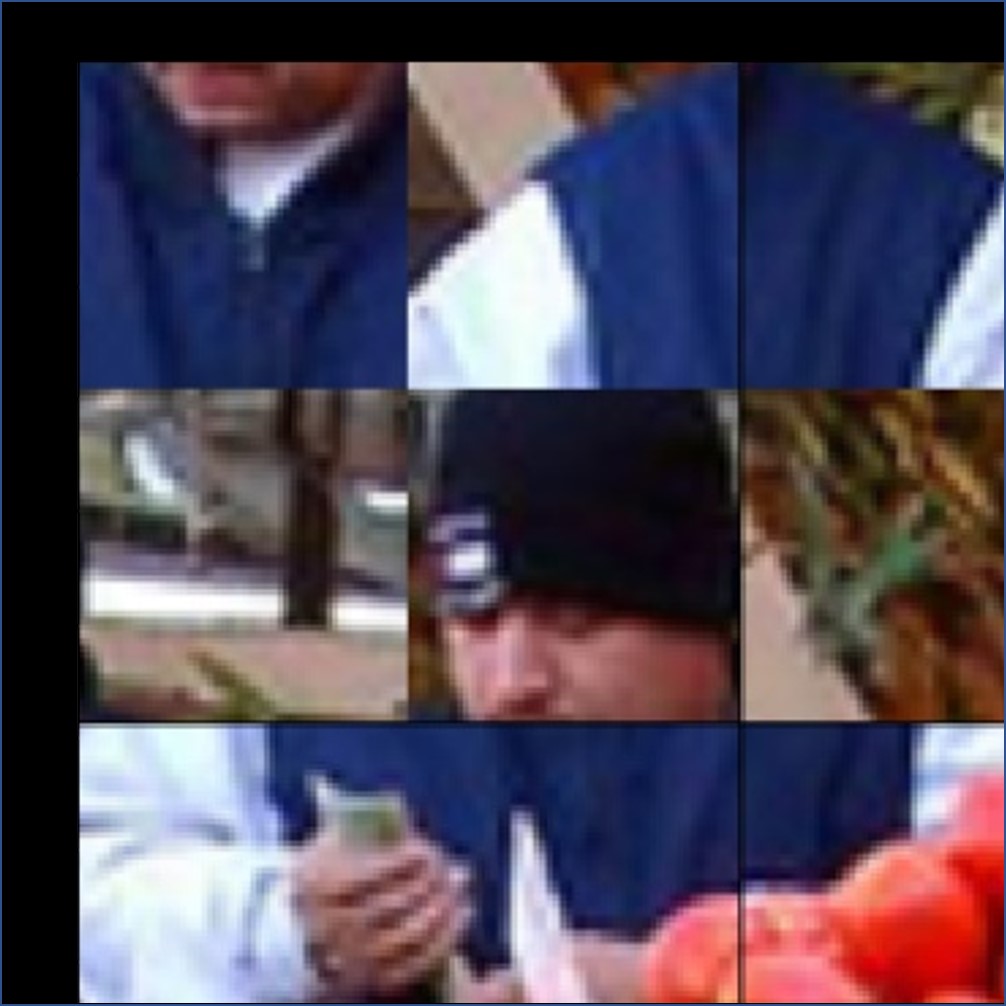}
}
\subfigure[Rotated]{
\includegraphics[width=.105\textwidth]{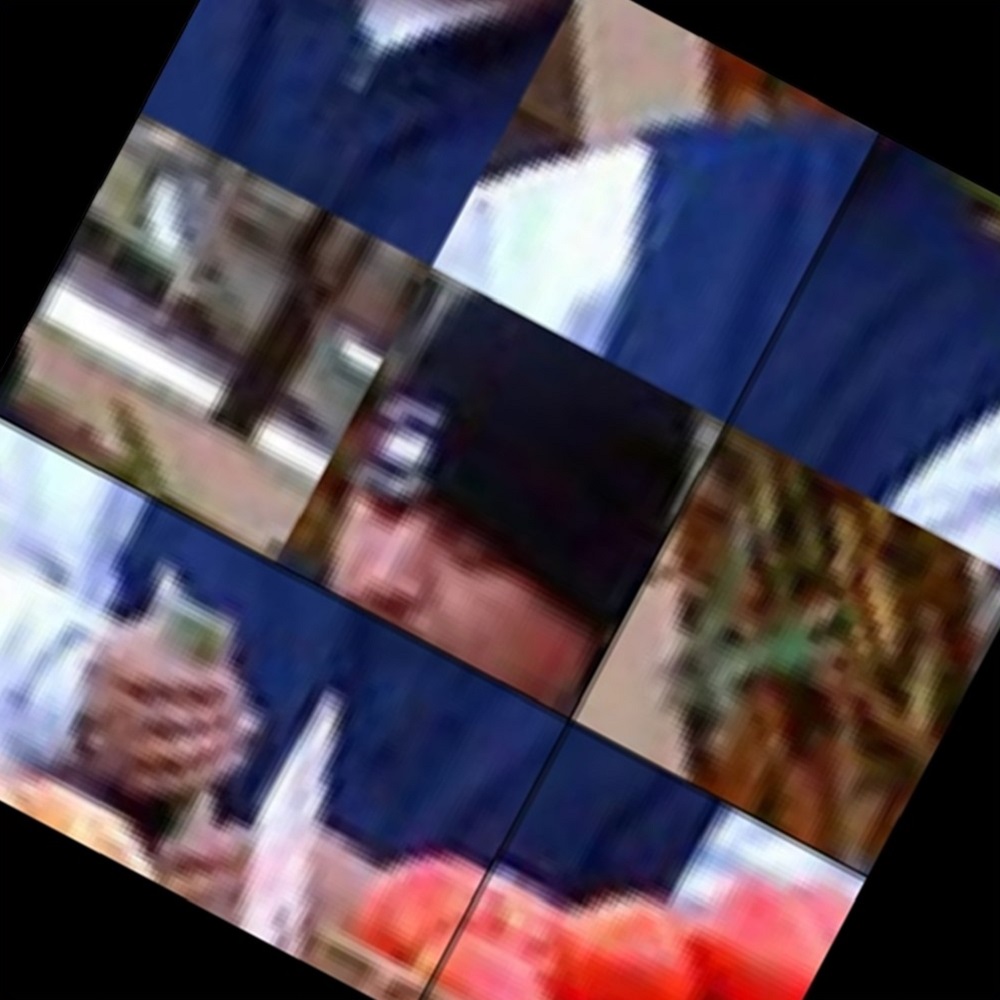}
}
\subfigure[Rescaled]{
\includegraphics[width=.105\textwidth]{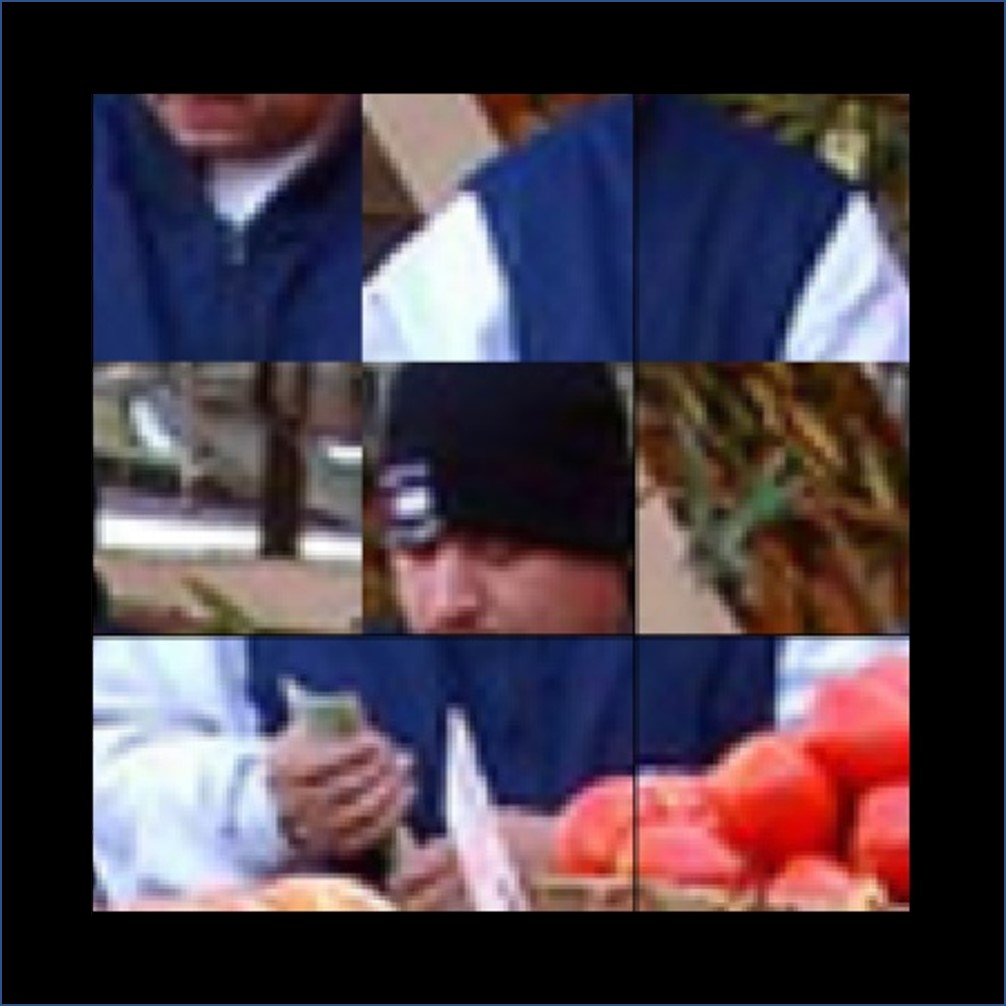}
}

\subfigure[Target Heatmap]{
\includegraphics[width=.105\textwidth]{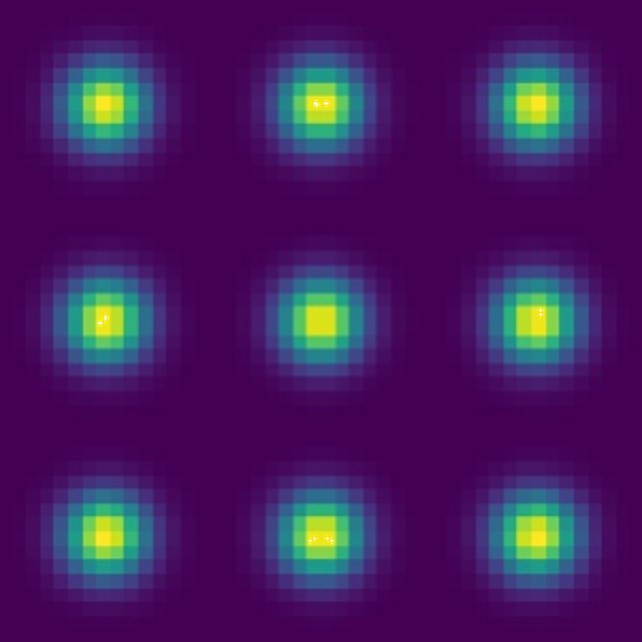}
}
\subfigure[Translated]{
\includegraphics[width=.105\textwidth]{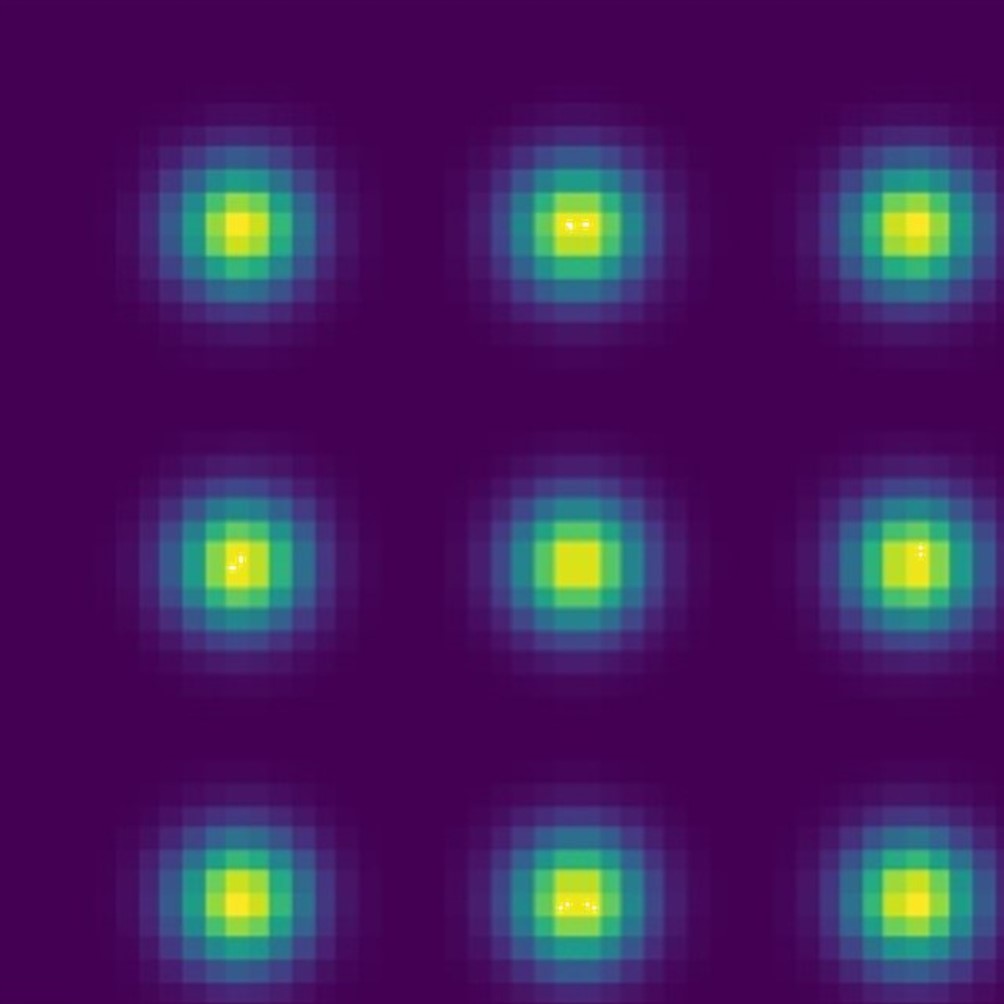}
}
\subfigure[Rotated]{
\includegraphics[width=.105\textwidth]{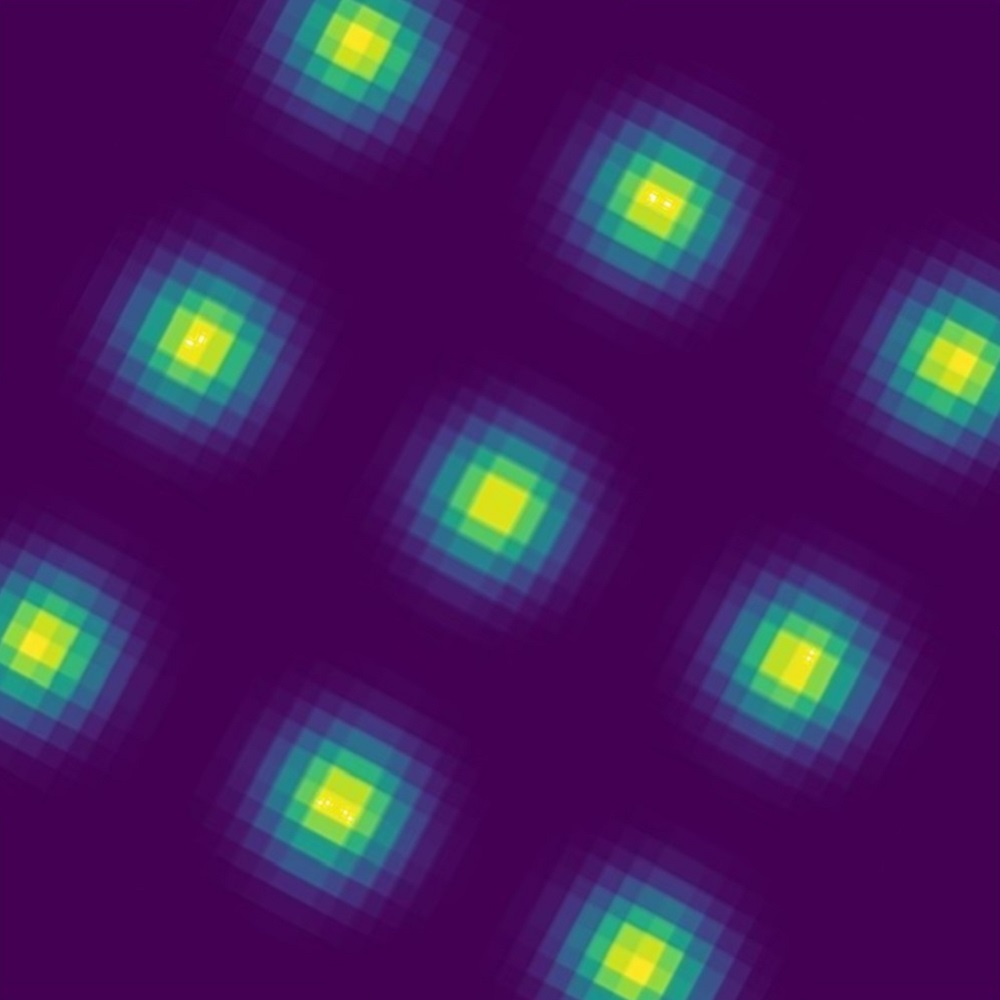}
}
\subfigure[Rescaled]{
\includegraphics[width=.105\textwidth]{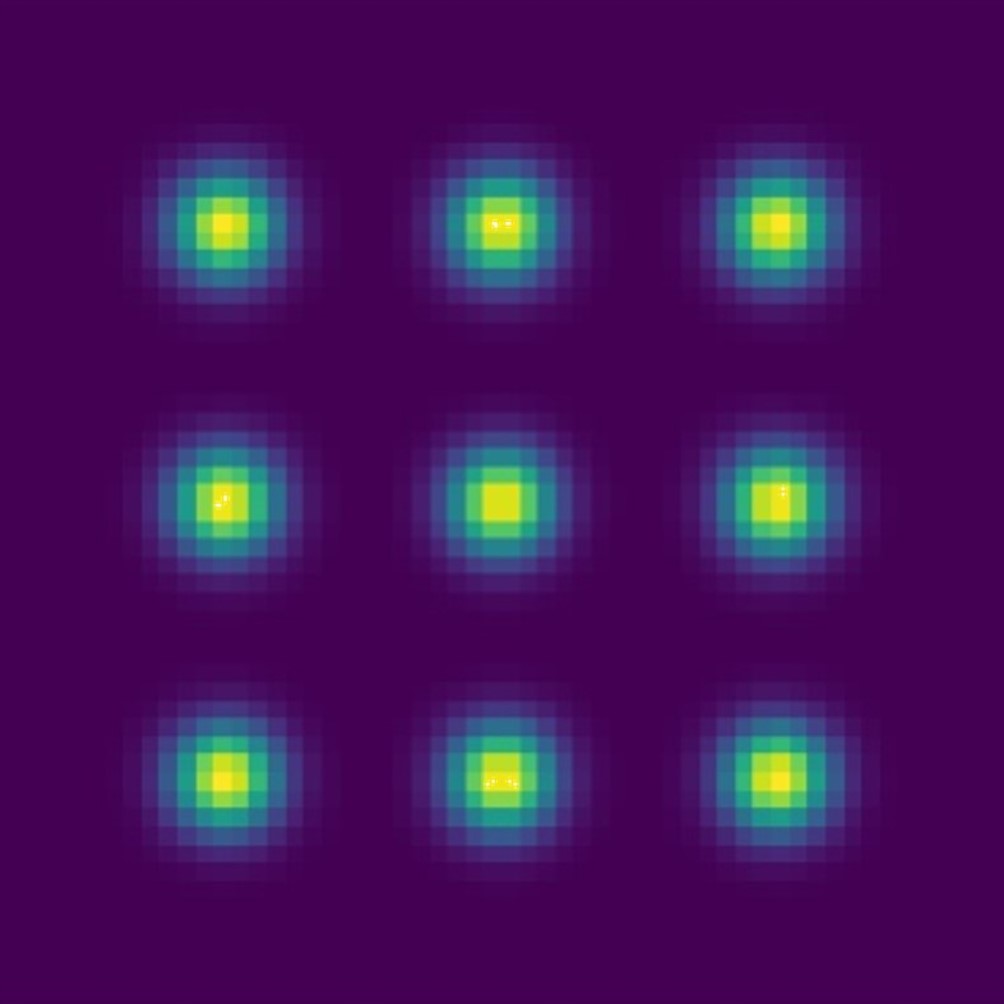}
}
\caption{Spatial data augmentations and possible heatmaps for all patches. In the first heatmap, without spatial data augmentations, the Gaussian Distributions are always on the centres of image grids, such as $(\frac{(2m+1)\cdot224}{2N},\frac{(2n+1)\cdot224}{2N})$. In the following three heatmaps, after rotating, rescaling, and translating augmentations in the training stage, the learnt responses are able to appear almost everywhere.} 
\label{spatialaug}
\end{figure}

There is obviously a challenging problem for our method proposed in Section \ref{tardis}: the centres of target 2D Gaussian Distributions are always located on centres of image grids, such as $(\frac{(2m+1)\cdot224}{2N},\frac{(2n+1)\cdot224}{2N}), m\in [0, N), n\in [0, N)$, so the network is easy to fix their responses on centres of grids (as is shown in the first heatmap of Figure \ref{spatialaug}) rather than having responses anywhere. In this case, it's hard for the network to learn rich representations for downstream tasks, because the learnt responses are highly related to output positions (whether they're grid centres), instead of the contents of image patches.

To tackle with this challenge, we introduce scaling $([-35\%, 35\%])$, translating $([-10\%, 10\%])$, and rotating $([-45^\circ, 45^\circ])$ augmentations randomly to the spatial concatenated shuffled images during the training process, as is demonstrated in the last three columns of Figure \ref{spatialaug}. The target positions of Gaussian Distributions corresponding to shuffled patch centres are also changed according to spatial augmentations. By this means, the union of responses learnt by certain neural network is possible to cover the entire 2-dimensional space of $224\times 224$ (except that $N$ is a small number like 1, 2), rather than obtaining responses only on centres of $N\times N$ grids like the first column in Figure \ref{spatialaug}. We don't care whether these spatial data augmentations are able to improve the performance of solving jigsaw puzzles, but they're beneficial to learn more spatial-comprehensive features for downstream tasks.

Apart from spatial augmentations, we follow common colour augmentations in recent papers of self-supervised learning \cite{simclr,boyl,mocov1} to improve the diversity of training dataset. Our colour augmentations for training include colour adjustments (hue, contrast, brightness, saturation), colour distortion, optional grayscale conversion, gaussian blur, and solarisation. By the means of colour and spatial augmentations, our HSJP task is able to provide the network with diverse semantics and comprehensive representations. 

\subsection{Loss function for HSJP pretext task}
As the form of our pretext task (HSJP) resemble its downstream task of 2D pose estimation - both of them predict some part of a person instance in a heatmap-based way, our loss function for HSJP pretext is similar to recent 2D pose estimation methods \cite{hrv1,simplebaseline,darkpose}. The output heatmaps from pose estimation networks \cite{hrv1,simplebaseline,darkpose} are usually $1/4$ the size of input images ($224\times 224$), and we follow the same settings: the output heatmaps for HSJP pretext task are also downsampled to $56\times 56$.

We utilise the same $\rm MSE$ loss function as \cite{hrv1,simplebaseline} to compare the predicted heatmap $\bm H$ with the groundtruth heatmap $\bm G$ during the training process. If the centre of $i$th patch is outside the image after spatial augmentation, the channel corresponding to $i$th patch is not considered into loss function during the training process. The $\rm MSE$ loss is calculated as follows:
\begin{equation}
{\rm MSE}(\bm G, \bm H)=\frac{1}{N^2\cdot 56^2}\sum_{i=1}^{N^2}\sum_{x=1}^{56}\sum_{y=1}^{56}\bm M_i\cdot(\bm G_{xy}^i-\bm H_{xy}^i)^2,
\end{equation}
where $\bm G_{xy}^i$ is the groundtruth heatmap response at location $(x, y)$ of channel $i$, and $\bm H_{xy}^i$ is the predicted response. $\bm M$ is a one-hot mask representing whether the centre of $i$th patch is inside the input image.

\subsection{Optimiser for HSJP pretraining}
Similar to common 2D pose estimation methods \cite{rsn,udppose,hrv1,simplebaseline,darkpose}, we employ the same Adam optimiser \cite{adam} to train the pretext task. Our models for HSJP are trained for 240 epochs on all person instances of COCO train2017 dataset with 4 Nvidia Quadro RTX 6000 GPUs, and our batch size is 256. The base learning rate is 1e-3, and it is dropped to 1e-4 and 1e-5 at the 190th and 220th epoch.

\begin{table*}
\begin{center}
{\caption{Comparing the performances of networks pretrained with HSJP, ImageNet, and training from scratch on COCO val2017 and test-dev2017 sets. We also report mAP of some other state-of-the-art networks \cite{cpn1,hourglass} in this table. (The performance without ImageNet pretraining is 0.734 in \cite{hrv1}, but we only reach 0.729 after several attempts.)}\label{apcmp}}
\begin{tabular}{l|c|c|c|c|c}
\hline
Approach&Backbone&Pretraining&Dataset&Input Size&AP\\
\hline
Stacked Hourglass \cite{hourglass}&8$\times$Stacked Hourglass&Scratch&val2017&$256\times192$&66.9\\
CPN + OHKM \cite{cpn1}&ResNet50 \cite{resnet}&ImageNet \cite{imgnet}&val2017&$256\times192$&69.4\\
\hline
SimpleBaseline \cite{simplebaseline}&ResNet50 \cite{resnet}&Scratch&val2017&$256\times192$&69.1\\
SimpleBaseline \cite{simplebaseline}&ResNet50 \cite{resnet}&ImageNet \cite{imgnet}&val2017&$256\times192$&70.4\\
SimpleBaseline \cite{simplebaseline}&ResNet50 \cite{resnet}&\textbf{HSJP, \textit{N}=6 (Ours)}&val2017&$256\times192$&\textbf{70.8}\\
\hline
HRNet W32 \cite{hrv1}&HRNet W32&Scratch&val2017&$256\times192$&72.9\\	
HRNet W32 \cite{hrv1}&HRNet W32&ImageNet \cite{imgnet}&val2017&$256\times192$&74.4\\	
HRNet W32 \cite{hrv1}&HRNet W32&\textbf{HSJP, \textit{N}=6 (Ours)}&val2017&$256\times192$&\textbf{74.5}\\
\hline
SimpleBaseline \cite{simplebaseline}&ResNet50 \cite{resnet}&Scratch&test-dev2017&$256\times192$&68.3\\
SimpleBaseline \cite{simplebaseline}&ResNet50 \cite{resnet}&ImageNet \cite{imgnet}&test-dev2017&$256\times192$&69.5\\
SimpleBaseline \cite{simplebaseline}&ResNet50 \cite{resnet}&\textbf{HSJP, \textit{N}=6 (Ours)}&test-dev2017&$256\times192$&\textbf{69.8}\\
\hline
HRNet W32 \cite{hrv1}&HRNet W32&Scratch&test-dev2017&$256\times192$&72.1\\	
HRNet W32 \cite{hrv1}&HRNet W32&ImageNet \cite{imgnet}&test-dev2017&$256\times192$&73.5\\	
HRNet W32 \cite{hrv1}&HRNet W32&\textbf{HSJP, \textit{N}=6 (Ours)}&test-dev2017&$256\times192$&\textbf{73.6}\\
\hline
\end{tabular}
\end{center}
\end{table*}

\begin{figure*}
\subfigure[(a) Finetuning on HRNet from Scratch.]{
\includegraphics[width=.32\textwidth]{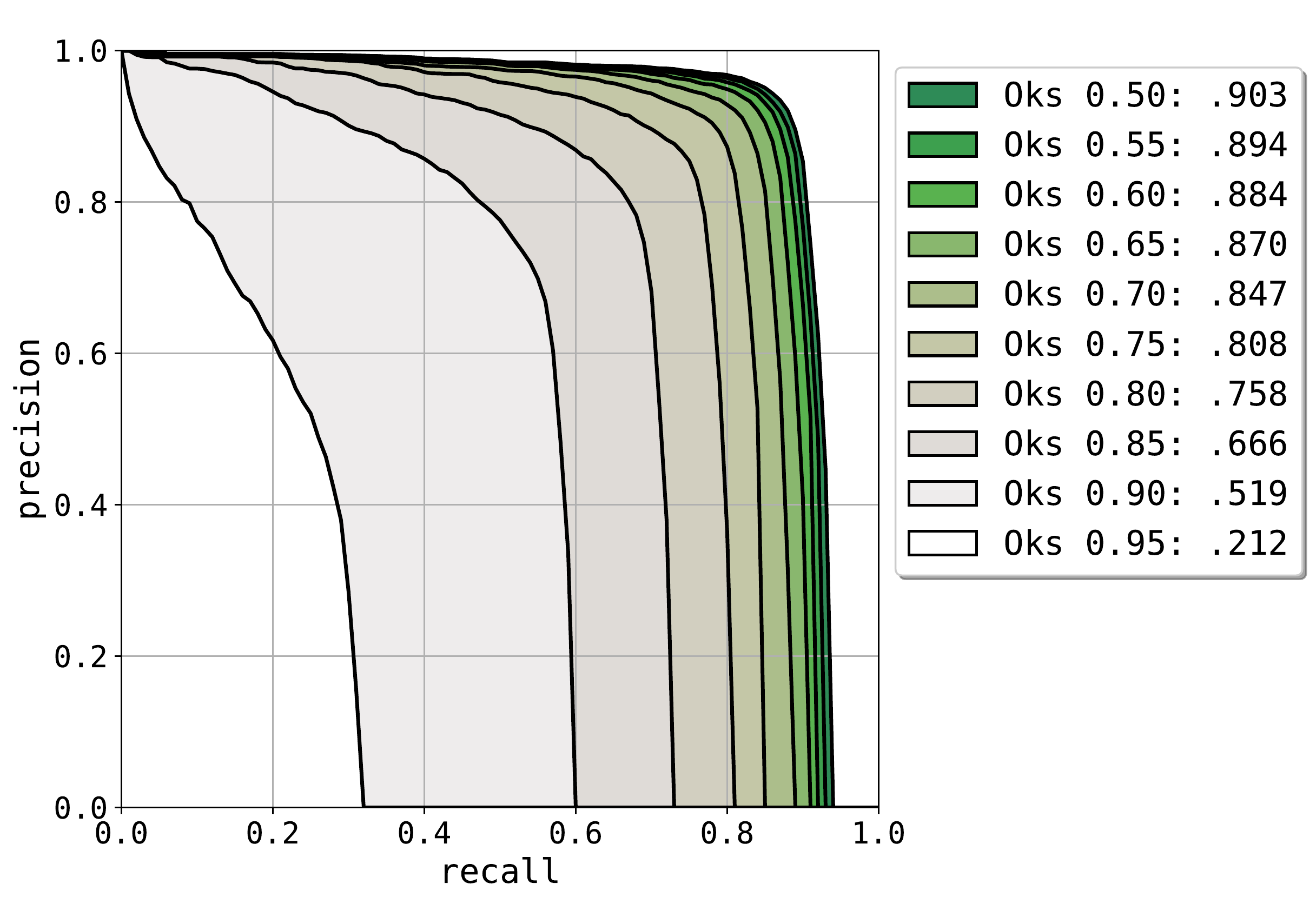}
\label{scratchdetail}
}
\subfigure[(b) Finetuning on HRNet from ImageNet.]{
\includegraphics[width=.32\textwidth]{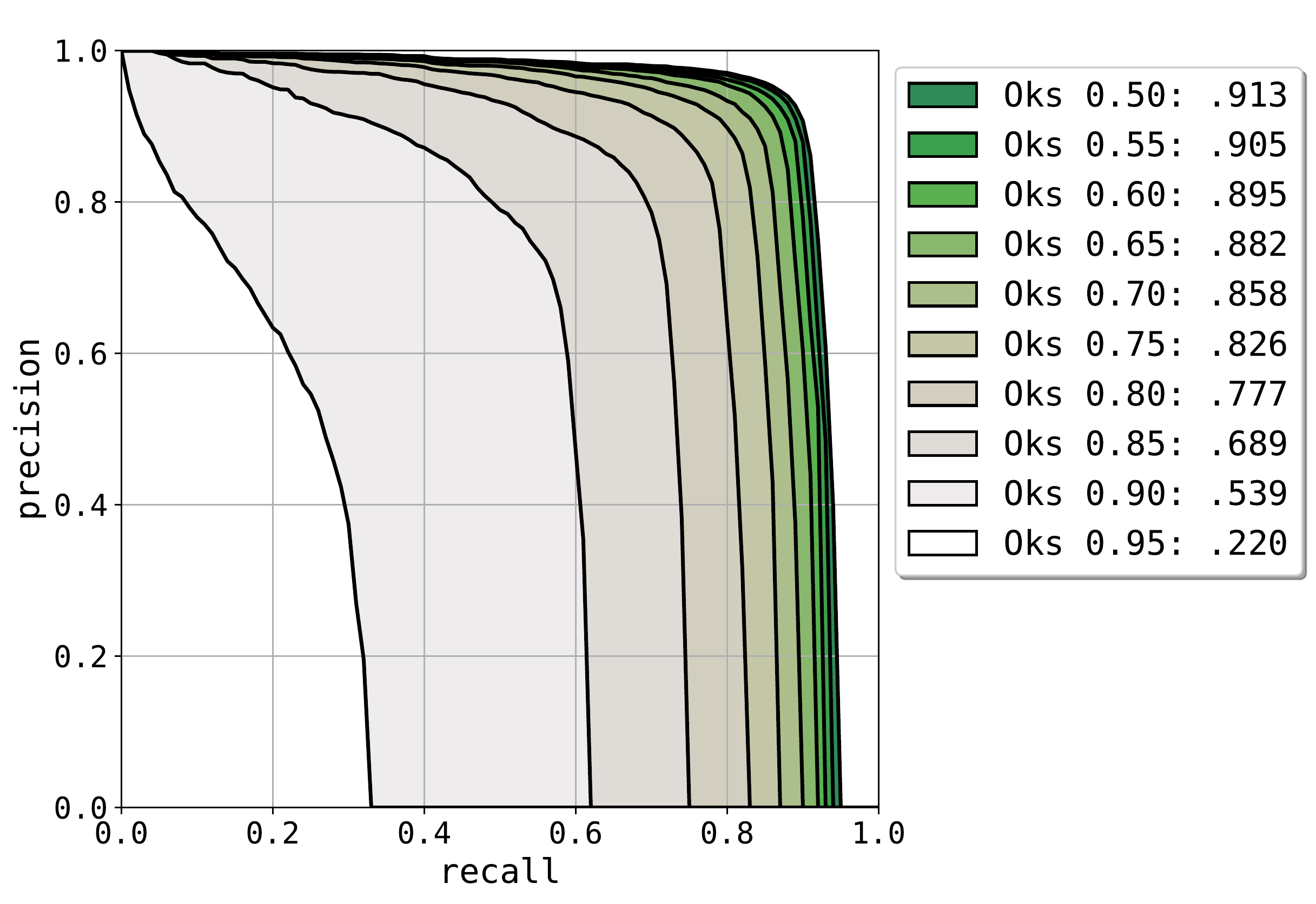}
\label{imgnetdetail}
}
\subfigure[(c) Finetuning on HRNet from HSJP.]{
\includegraphics[width=.32\textwidth]{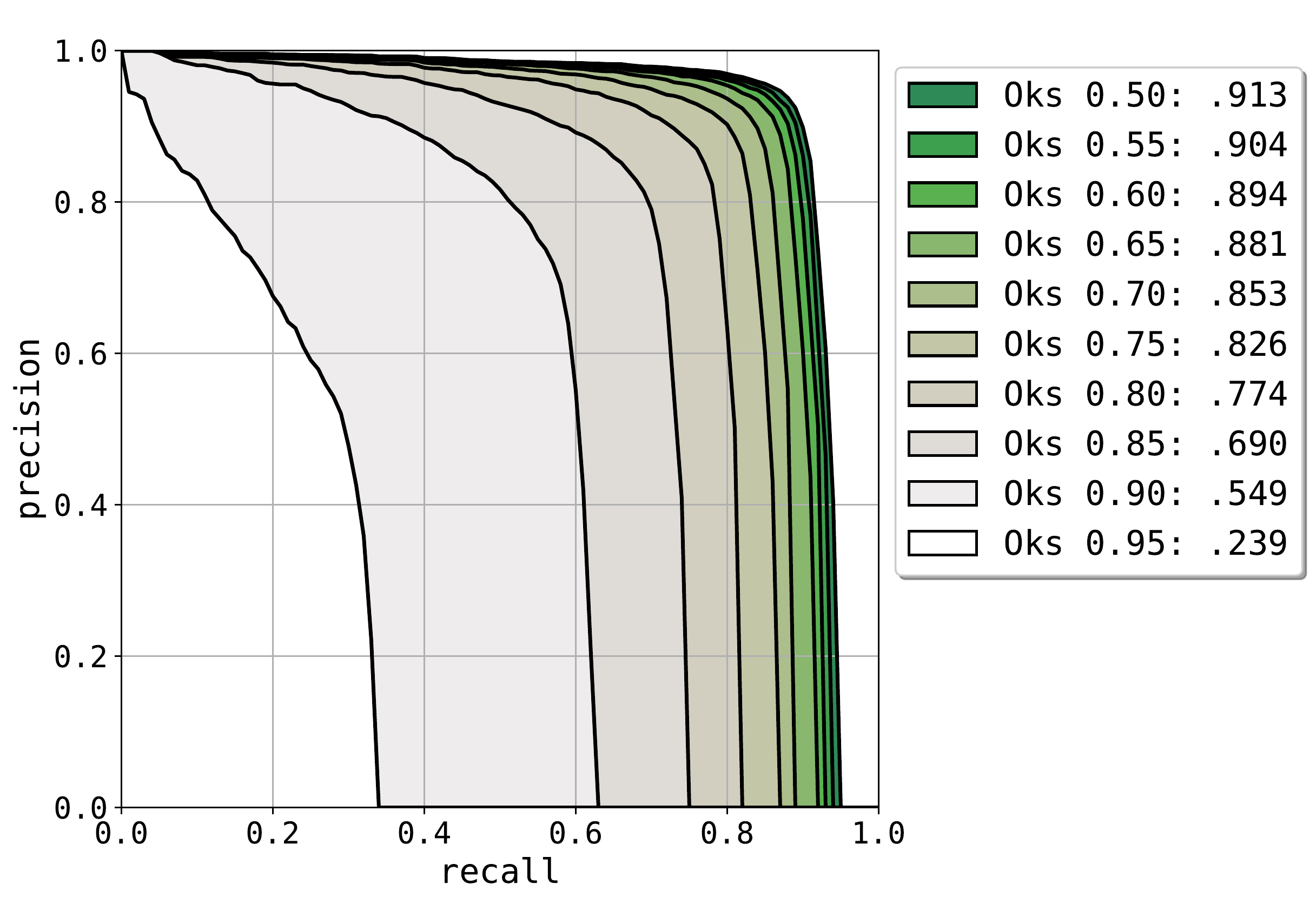}
\label{hsjpdetail}
}
\caption{Detailed performance analysis of downstream task, based on HRNet W32 pretrained with different approaches. Both ImageNet and HSJP pretraining perform much better than training from scratch in their downstream task, for all OKS thresholds. HSJP share similar performance with ImageNet pretraining at low OKS thresholds (ImageNet slightly better), while HSJP pretrained HRNet performs much better for 2D human pose estimation at high thresholds (OKS $\geq$ 0.85).} 
\label{detailedperformance}
\end{figure*}

\subsection{Performance of solving jigsaw puzzles}
\label{jigsaweval}
To estimate the representation power of CNN after pretraining on HSJP pretext, we evaluate the average precision of jigsaw puzzles on all person instances in COCO val2017 dataset. We split and shuffle image patches as Section \ref{tardis} does, but we don't use any data augmentation when testing the performance of HSJP, because the randomness of test-time augmentation makes the performance unstable. For the $i$th patch, if the peak of predicted Gaussian Distribution $\hat{\bm p}_i$ lies in the neighbourhood of its corresponding ground-truth patch centre $\bm p_i$, i.e.,
\begin{equation}
||\hat{\bm p}_i-\bm p_i||_2<\epsilon,
\end{equation}
we consider that the $i$th image patch is an accurately located one for solving jigsaw puzzles ($||\cdot||_2$ represents the L2 norm). If all of the $N\times N$ patches in a shuffled image are correctly located, i.e.,
\begin{equation}
||\hat{\bm p}_i-\bm p_i||_2<\epsilon,\forall i\in [0,N^2)
\end{equation}
the jigsaw puzzles problem for that image is considered as successfully solved. 

We report the average precision of HSJP for all person instances in COCO val2017 dataset, i.e. 
\begin{equation}
P=\frac{T}{T+F}
\end{equation}
where $T$ represent the number of correctly solved jigsaw puzzles across all person instances in COCO val2017, and $F$ is the number of unsuccessfully solved ones. During our pretraining, the indicator $P$ is used to select a powerful network for solving HSJP problem, whose weights are then utilised for initialising downstream 2D pose estimators.

\section{Experiments}
This section is mainly composed of ablation studies and the downstream performances of 2D human pose estimation on MS-COCO \cite{coco} val2017 and test-dev2017 datasets.
\label{exp}

\subsection{Dataset and metric for downstream task}
The MS-COCO human keypoint dataset includes more than 200k images and 250k person instances, and 150k of them are publicly available for training. We utilise OKS-based Mean Average Precision \cite{coco} (mAP score) under 10 different thresholds as evaluation metric on MS-COCO dataset, where OKS  calculates the similarities between predicted keypoints and ground truth positions. 

\subsection{Finetuning on 2D human keypoint labels}
\label{cocofine}
We pretrain networks with HSJP on all person instances in MS-COCO train2017 dataset. After pretraining, we finetune our learnt weights on human keypoint labels with two popular 2D pose estimators, HRNet \cite{hrv1} and  SimpleBaseline \cite{simplebaseline}. For fair comparison, we follow their training settings. The data augmentations include random rotation $([-45^\circ, 45^\circ])$, rescaling $([-35, 35])$, and flipping. We also maintain the same number of training epochs as \cite{simplebaseline,hrv1} under the same Adam optimiser, and the input size of our downstream task is fixed on $256\times 192$. The only difference in our experiment is whether we initialise the downstream network with weights pretrained on labelled ImageNet \cite{imgnet} dataset or on our self-supervised HSJP pretext task. For fair comparison, we also follow the same two-stage paradigm in SimpleBaseline \cite{simplebaseline}, and HRNet \cite{hrv1}: we firstly detect all person instances with Faster-RCNN \cite{fastercnn} and then estimate keypoints of detected human bodies. 

The performances of different pretraining methods and networks are reported in Table \ref{apcmp}, which shows that the downstream 2D human pose estimators with our HSJP pretraining achieve much better performance than training from scratch, and the results are comparable to ImageNet pretraining (even a little better than ImageNet for 0.x\%). 

\subsection{Detailed analysis of downstream performance}

\begin{figure}
\subfigure[From Scratch]{
\includegraphics[width=.146\textwidth]{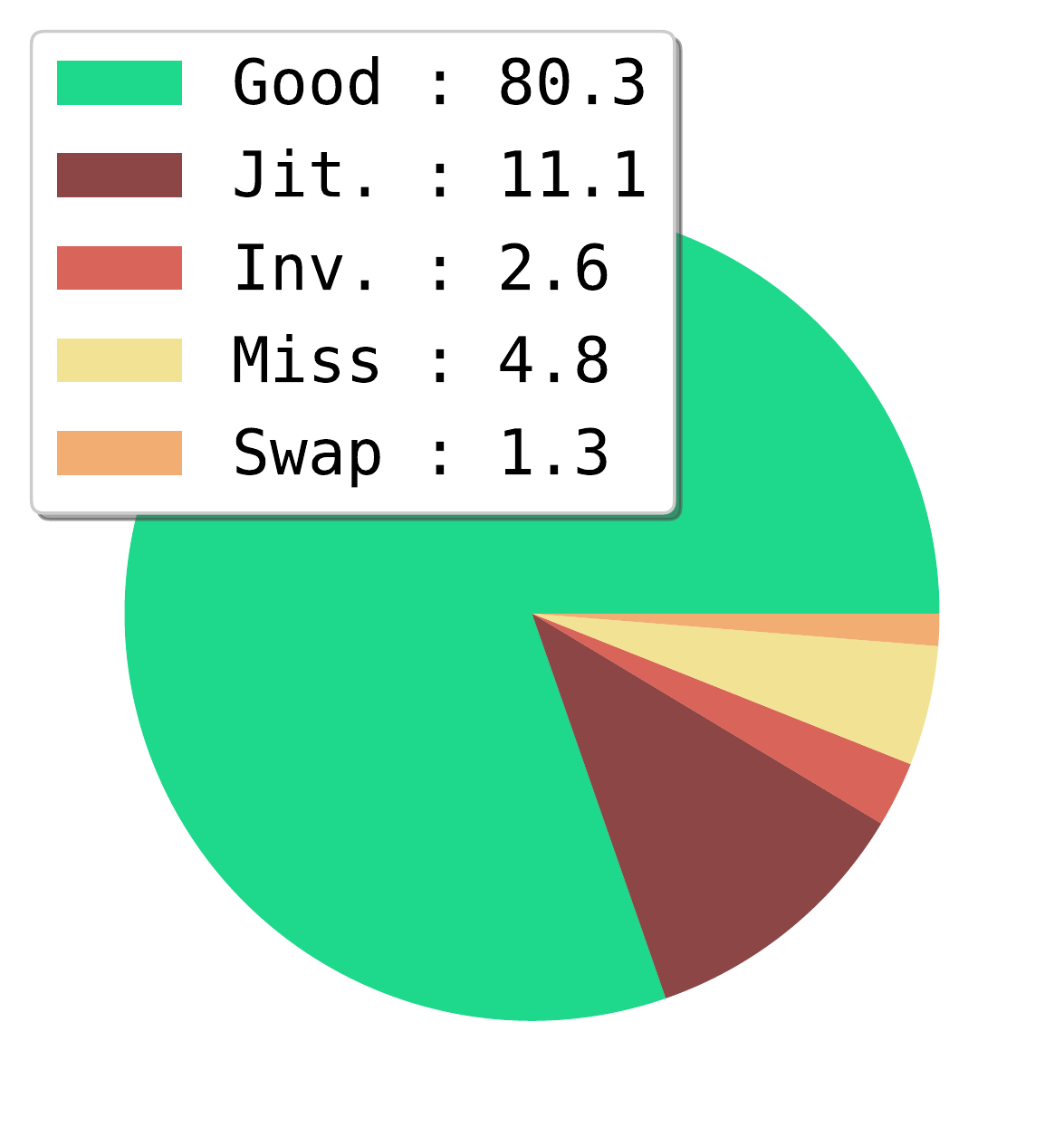}
}
\subfigure[From ImageNet]{
\includegraphics[width=.146\textwidth]{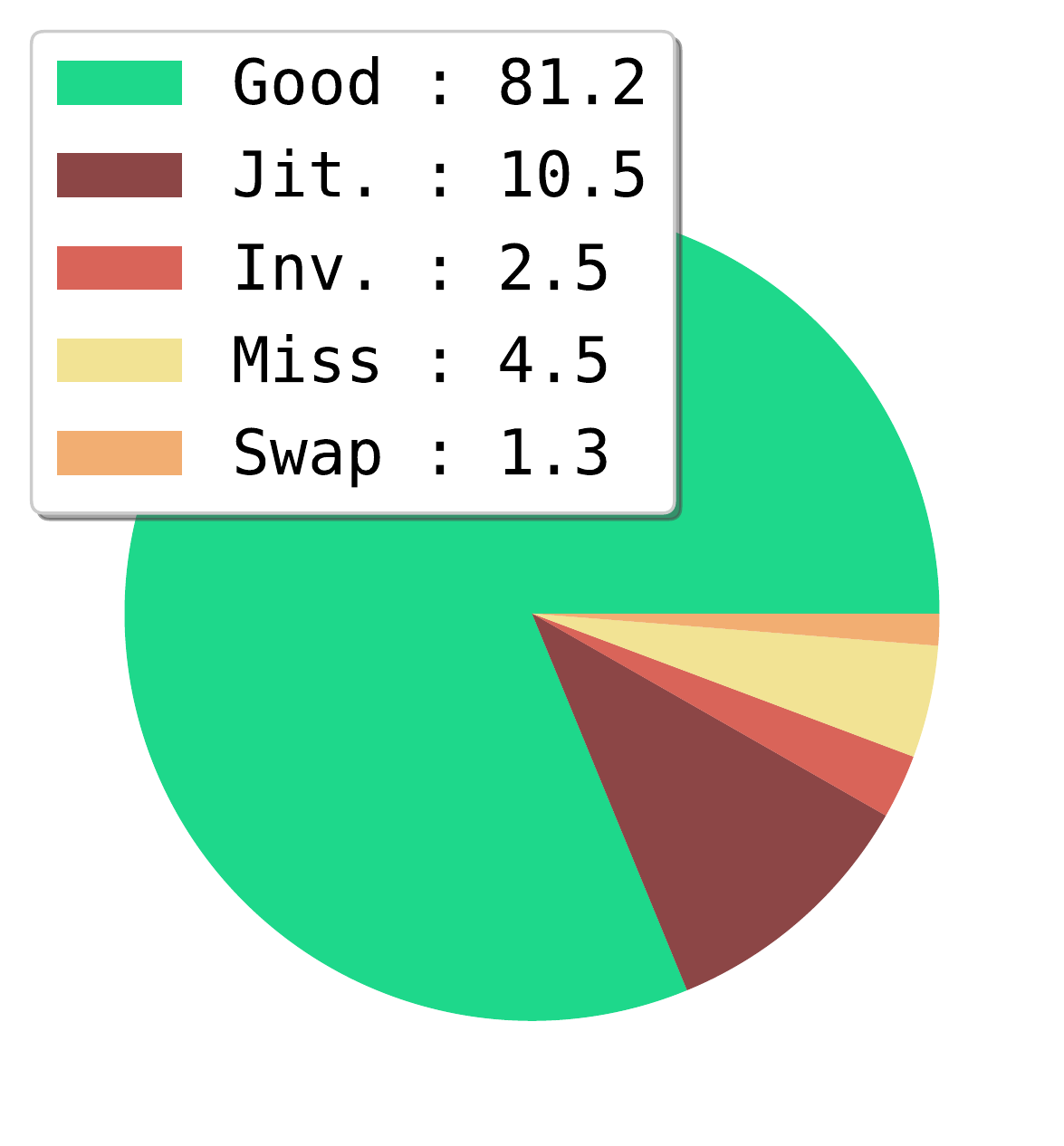}
}
\subfigure[From HSJP]{
\includegraphics[width=.146\textwidth]{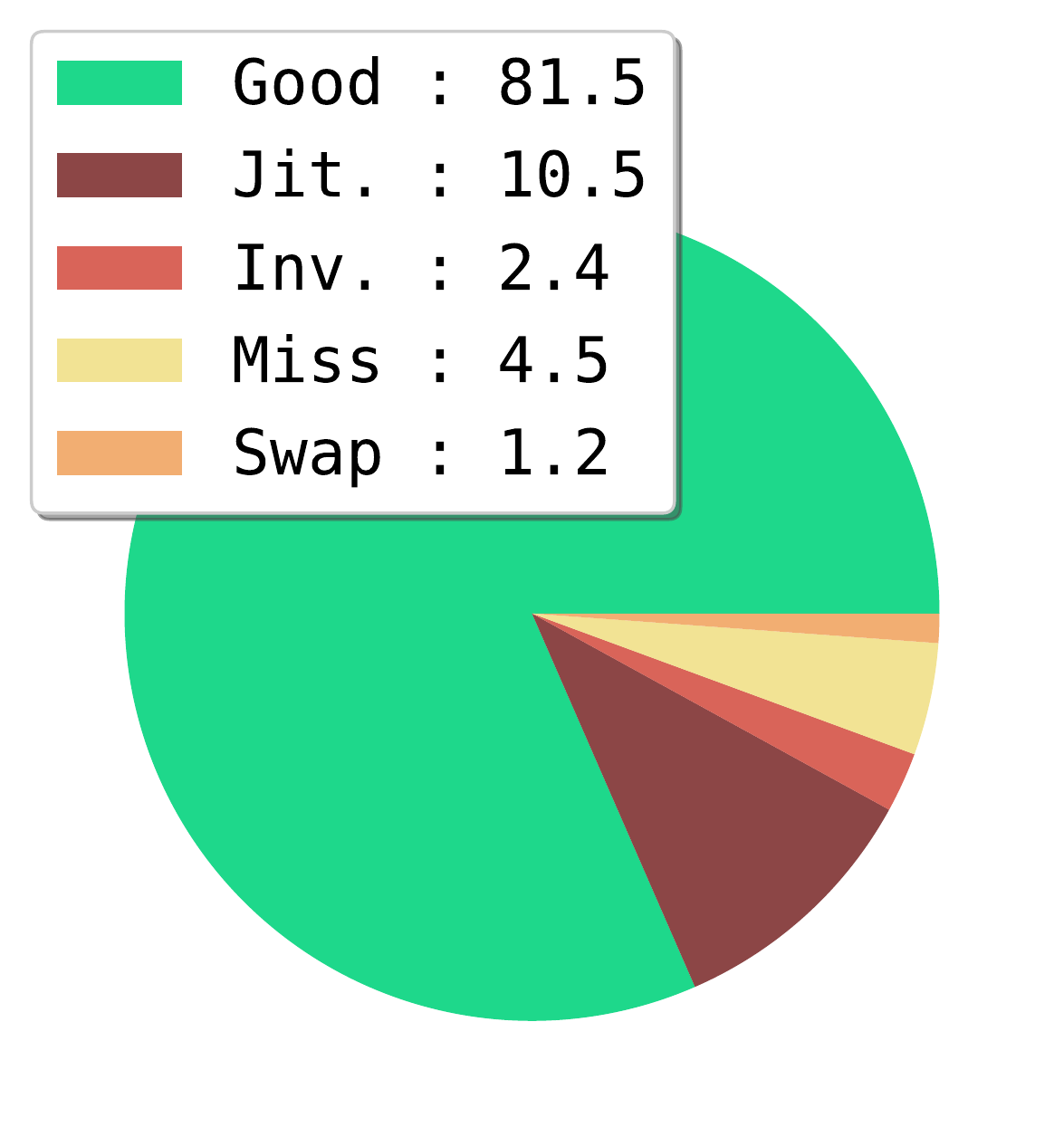}
}
\caption{Comparing errors of 2D pose estimation, via HRNet W32 with different pretraining methods. Our HSJP pretraining method enjoys the best downstream performance.} 
\label{erroranalysis}
\end{figure}

\begin{figure}
\includegraphics[width=.48\textwidth]{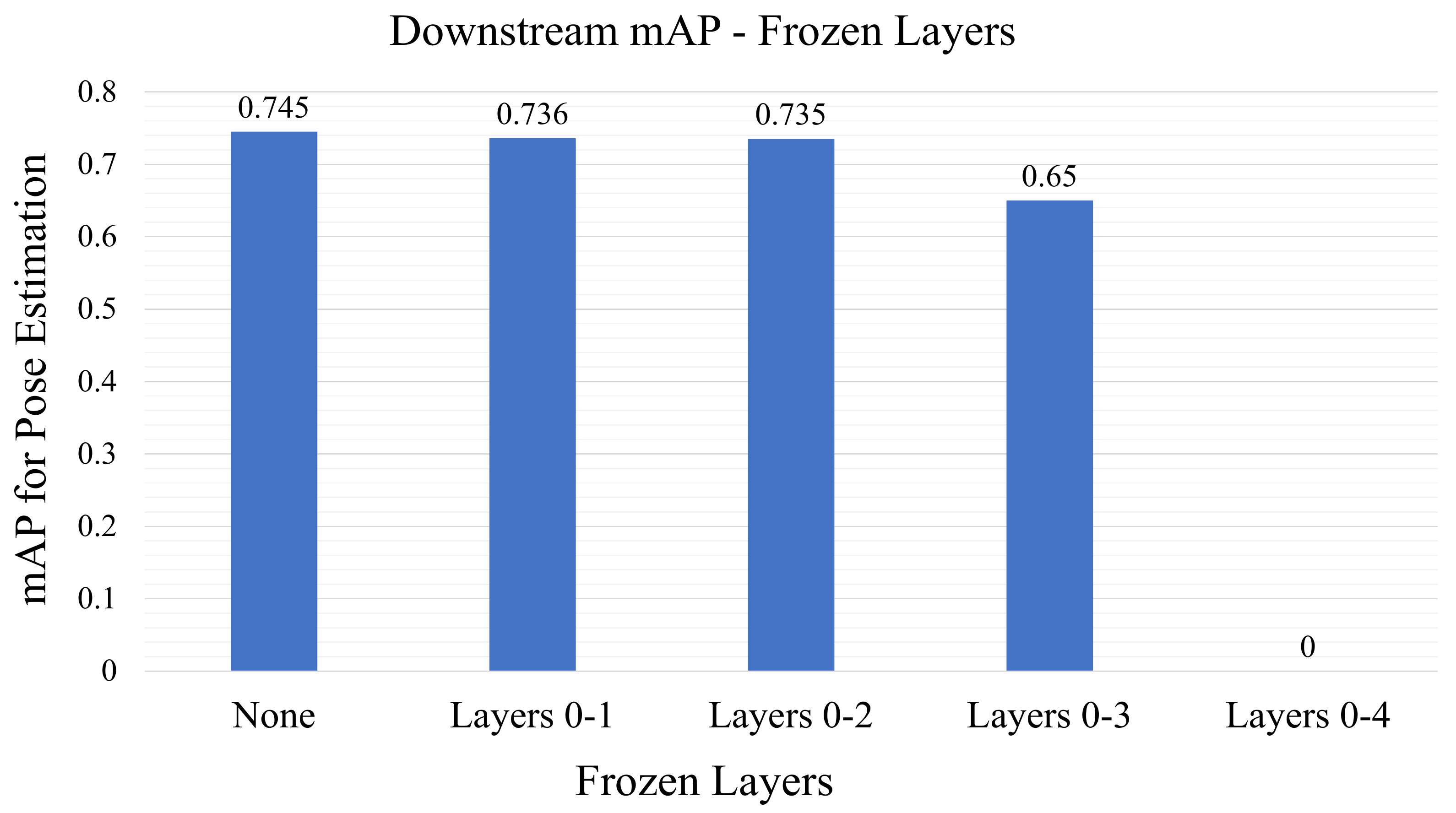}
\caption{Analysing the transferability of HRNet W32 pretrained with HSJP pretext task. We freeze different layers of HRNet W32 during finetuning and report their mAP for 2D pose, to explore which layers are general or task-specific.\label{trans}} 
\end{figure}

We also analyse detailed downstream performances with different OKS thresholds (from 0.5 to 0.95) on HRNet W32, as Figure \ref{detailedperformance} shows. It's easy to conclude that pretrained HRNet W32 (with ImageNet or HSJP) achieve much better precision at all OKS thresholds than training from scratch. Comparing the performance of ImageNet with HSJP pretraining in Figure \ref{imgnetdetail} and \ref{hsjpdetail}, we can see that ImageNet pretrained HRNet performs slightly better (like 0.1\%) at low OKS thresholds (for OKS $\leq$ 0.8), while HSJP pretrained HRNet W32 performs much better ($\geq$ 1\%) at high thresholds (for OKS $\textgreater$ 0.8). Although ImageNet and HSJP share similar mAP on 2D human pose estimation, ImageNet pretrained HRNet W32 achieve this mAP score by slightly accumulating improved low-threshold performance, while HSJP pretrained HRNet W32 predicts more accurately at high OKS thresholds for 2D human pose estimation.

Also, we use the method in \cite{erroranalysis} to analyse the error of HRNet W32 under different pretraining methods, as Figure \ref{erroranalysis} demonstrates, and we can draw a similar conclusion. Both ImageNet and HSJP pretrained HRNet perform much better than training from scratch. HSJP pretrained HRNet W32 enjoys the highest ratio of good prediction, and the lowest ratio for inverting/swapping errors. ImageNet and our HSJP pretraining perform similarly for jitter and missing errors.

Taking the finetuning experiments into consideration, we conclude that HSJP pretrained 2D human pose estimators are able to achieve much better performance than those trained from scratch, and their mAP score on the downstream task are also comparable to ImageNet pretrained networks. However, we don't use any supervised data in our HSJP pretraining, which saves the labour costs of labelling extra datasets (such as ImageNet) for pretraining.

\subsection{Transferability analysis of different layers}

\begin{figure}
\centering
\includegraphics[width=.47\textwidth]{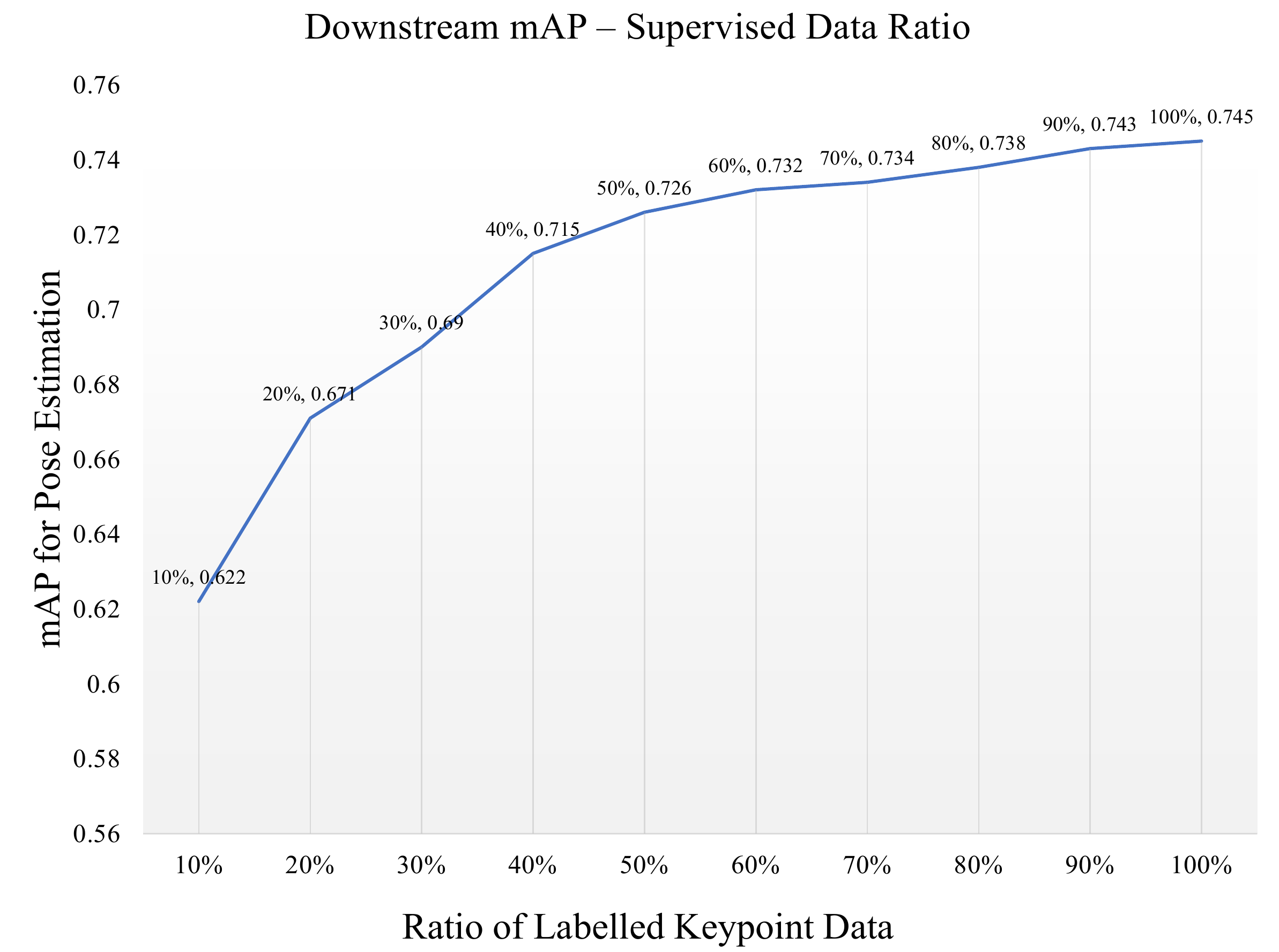}
\caption{Semi-supervised training with 10\% to 90\% of MS-COCO human keypoint labels.\label{semi}} 
\end{figure}

To explore which layers are general or task-specific, we perform similar experiment as \cite{jigsaw16}: we freeze some layers of HSJP pretrained HRNet W32 during finetuning, to test the transferability of learnt feature maps. Our analyses on 4 layers of HRNet W32 are reported in Figure \ref{trans} as follows.

(1) When freezing the top layers (0 to 2) of HSJP pretrained HRNet W32 during finetuning, the mAP for downstream 2D pose estimator does not decrease so much, so the top layers learnt by HSJP pretext are general for extracting image features.
(2) When freezing layers 0 to 3 of HSJP pretrained HRNet in the downstream task, the mAP score for 2D pose estimation decrease a lot, but the network does not collapse. Therefore, some features in layer 3 are general for feature extraction, and some features need to be task-specific for the downstream task. (3) When we freeze all weights in layers 0 to 4 of HRNet W32 learnt by HSJP pretext task, just leaving the last feature fusion layer for final heatmap regression, the downstream performance collapses to mAP of 0. Hence we know that the layers next to the output of network are specific for downstream task.

\subsection{Semi-supervised training on keypoint labels}

The HSJP pretext task is able to build a semi-supervised scheme for 2D human pose estimation: after pretraining backbone network with HSJP pretext task, we fine-tune the HSJP pre-trained network on a randomly selected subset (such as 10\%, 20\%) of the MS-COCO data set. We do not use any human keypoint labels during pretraining, and we only use partial of 2D human keypoint labels in MS-COCO dataset during finetuning. In this way, our method is semi-supervised. In our experiments for semi-supervised 2D pose estimation, we finetune the HSJP pretrained HRNet W32 on several subsets (from 10\% to 90\%) of MS-COCO human keypoint dataset, and we finally demonstrate the performances (mAP for 2D human pose estimation) of our semi-supervised learning in Figure \ref{semi}.

\section{Ablation Studies}
\label{ablation}
We finally discuss several hyperparameter settings in this section. As we follow the previous settings of 2D human pose estimation \cite{simplebaseline,hrv1} in the downstream task, our main concern is about hyperparameters in HSJP pretext task. The ablation studies mainly include three parts: (1) learning rate and number of epochs; (2) how many patches ($N\times N$) should we divide the original input image into; (3) whether using initial (unshuffled) image in HSJP pretext task. 

\begin{table}
\begin{center}
{\caption{Ablation studies on epoch numbers and learning rates for training HSJP pretext task. The experiments are performed with HRNet W32. The best performance with least cost comes from learning rate of 1e-3 and 240 epochs.}\label{epochlr}}
\subtable[Performances of different learning rates.]{
\label{lrablation}
\begin{tabular}{c|c|c}
\hline
Learning Rates&HSJP Precision&mAP for 2D Pose\\
\hline
3e-4&0.355&0.742\\
\textbf{1e-3}&\textbf{0.368}&\textbf{0.745}\\
2e-3&0.0(Collapse)&Not Available\\
\hline
\end{tabular}
}
\subtable[Performances of different epoch numbers.]{
\begin{tabular}{c|c|c}
\hline
Epoch Numbers&HSJP Precision&mAP for 2D Pose\\
\hline
120&0.297&0.739\\
\textbf{240}&\textbf{0.368}&\textbf{0.745}\\
480&0.407&0.745\\
\hline
\end{tabular}
}
\end{center}
\end{table}

\subsection{How many epochs under which learning rate?}
Our ablation study in this subsection is to explore the best epoch number and learning rate for training HSJP pretext task. We try 3 different learning rates (3e-4, 1e-3, and 2e-3) as well as 3 different epoch numbers (120, 240, and 480) for our HSJP pretext task on HRNet W32, and their performances are reported in Table \ref{epochlr}. When training HSJP pretext task for 120 epochs, the network cannot learn rich enough representations, and the downstream task cannot reach its best performance. When training 480 epochs, the precision on HSJP pretext is higher than that of 240 epochs, but it does not help to increase the mAP of downstream task: 2D human pose estimation. We finally select 240 as the number of epochs for training our HSJP pretext task, as it's not only of high performance but also time-saving (Training for 480 epochs is time-consuming, but useless for downstream task). Table \ref{lrablation} illustrate that the pre-task can converge quickly at learning rate of 1e-3, and the network does not collapse. Thus we select 1e-3 as our learning rate.

\begin{table}
\begin{center}
{\caption{Ablation studies on splitting number $N$ and whether using unshuffled image. We can conclude that training without original image, and $N=6$ provides the best performance for downstream 2D pose estimation. The setting "From Scratch (450)" in our table trains HRNet for 450 epochs to balance the HSJP pretraining epochs (210+240) for fair comparison. }\label{cmpcontrast}}
\subtable[HRNet W32 Performances pretrained without unshuffled image]{\label{splitablation}
\begin{tabular}{c|c|c}
\hline
$N$ for Splitting&HSJP Precision&mAP for Pose\\
\hline
2&0.993&0.739\\
3&0.954&0.739\\
4&0.766&0.742\\
5&0.533&0.743\\
\textbf{6}&\textbf{0.368}&\textbf{0.745}\\
7&0.243&0.743\\
8&0.019&0.735\\
\hline
From Scratch (210)&Not Available&0.729\\
From Scratch (450)&Not Available&0.734\\
\hline
\end{tabular}
}
\subtable[HRNet W32 Performances pretrained with original unshuffled image]{
\begin{tabular}{c|c|c}
\hline
$N$ for Splitting&HSJP Precision&mAP for Pose\\
\hline
2&0.995&0.734\\
3&0.989&0.734\\
4&0.972&0.735\\
5&0.954&0.735\\
6&0.936&0.734\\
7&0.892&0.735\\
8&0.777&0.735\\
\hline
From Scratch (210)&Not Available&0.729\\
From Scratch (450)&Not Available&0.734\\
\hline
\end{tabular}
}
\end{center}
\end{table}

\subsection{How many pieces of image patches?}
\label{patchesub}

We need to decide how many patches (the number $N$) should we divide the original image into during HSJP pretraining. If the number of image patches is too small (like $N=1,2$), the network will be easy to converge, so it will be hard for the network to learn feature maps with rich representations, as the jigsaw puzzling task is too simple. However, if the number of image patches is too large (such as $N>8$), the jigsaw puzzling task is too complicated, and  as our resolution of input image is limited to $224\times 224$, the pretraining procedure may not be easy to converge. 

For this concern, we test different settings of $N$ for our HSJP pretext task with HRNet W32 in this subsection, and finetune networks pretrained with different $N$ on the downstream 2D pose estimator. The HSJP and downstream mAP performances on MS-COCO val2017 dataset are reported with the metrics in Section \ref{jigsaweval} and \ref{cocofine}. As Table \ref{splitablation} shows, with the increment of $N$, the mAP score for downstream 2D human pose estimator firstly increases, and then decreases. However, the HSJP precision decreases with the increment of $N$, because the complexity of jigsaw puzzling problem also increases. When $N=6$, the convergence and the representation capability of pretext task are able to reach a balance: the pretraining process on HSJP pretext task does not collapse, and the network also learns high-quality representations for downstream 2D pose estimator (although it's of low HSJP precision, our purpose is not to solve jigsaw puzzles more accurately, but to learn representations for downstream task). Therefore, we split our original input image of $224\times 224$ into $6\times 6$  image patches.

\subsection{Whether using unshuffled image in pretext?}

\begin{figure}
\subfigure[(a) Using Shuffled Patches Only.]{
\raisebox{.4\height}{\includegraphics[width=.225\textwidth]{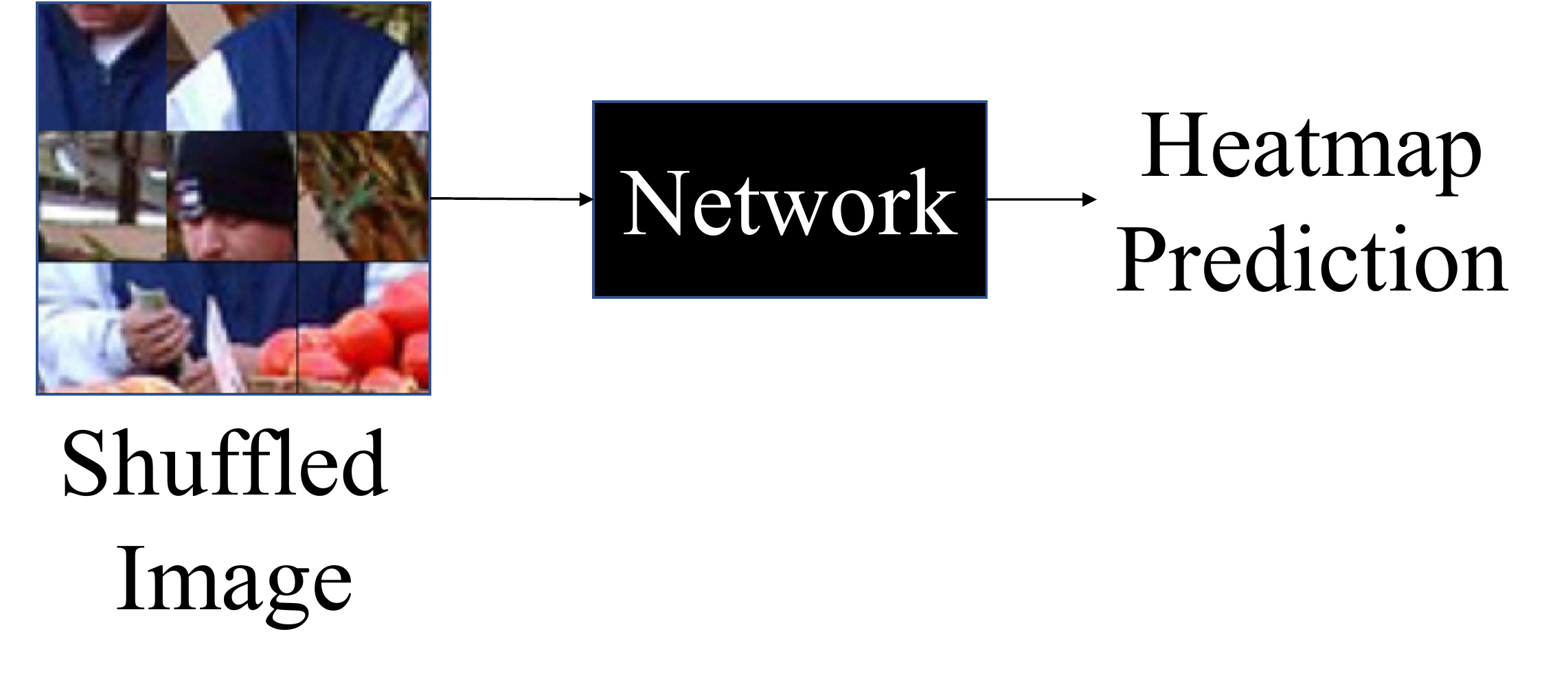}}
}
\subfigure[(b) Adding Original Person Image.]{
\includegraphics[width=.225\textwidth]{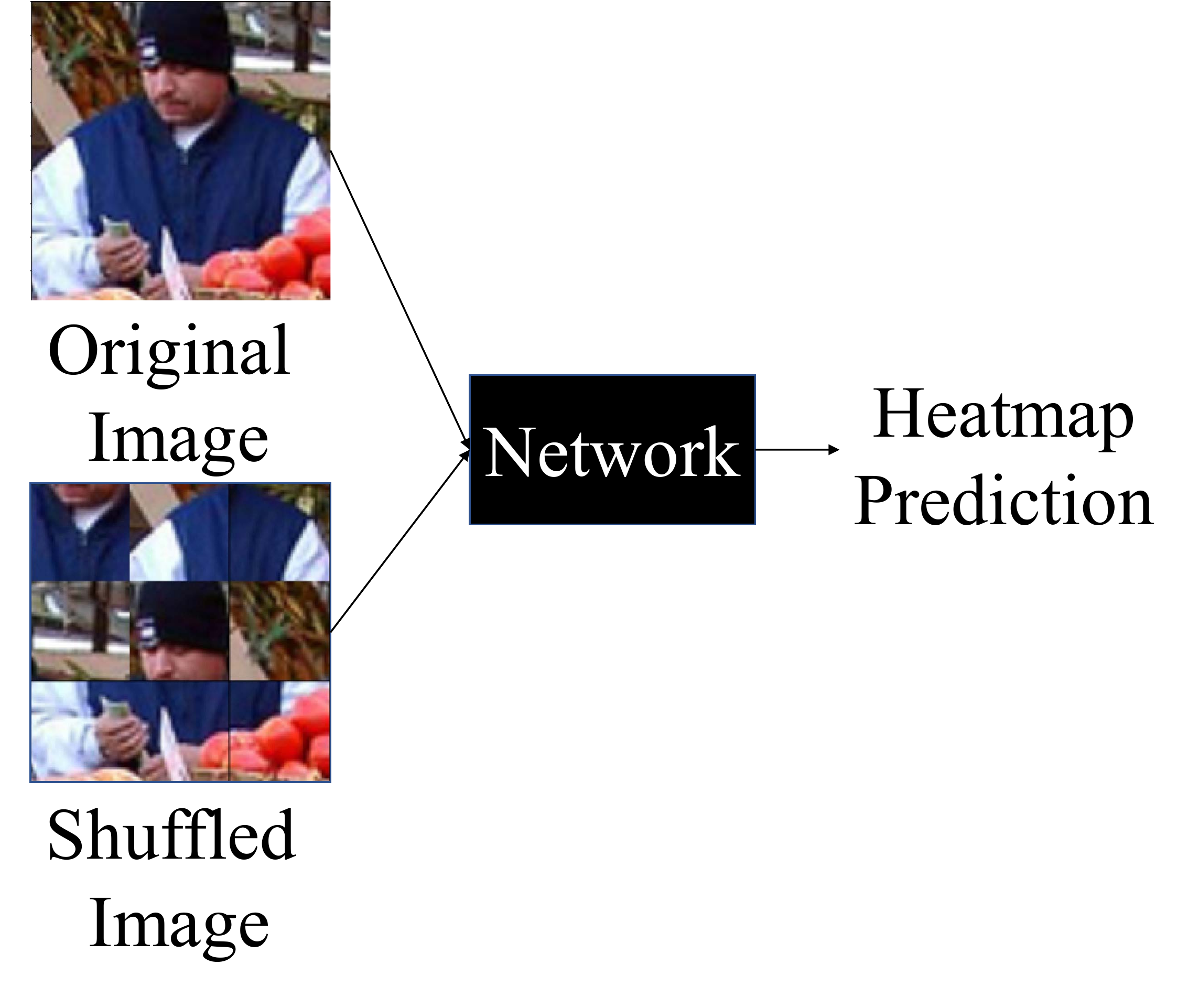}
}
\caption{Whether using original person image in HSJP.\label{addorigin}} 
\end{figure}

Similar to \cite{jigsaw19}, we consider concatenating original unshuffled person image to the input of HSJP pretext, as is shown in Figure \ref{addorigin}, for two reasons: (1) Taking unshuffled image as input makes the pretraining process similar to the downstream task of 2D human pose estimation, which also takes original person instance as input; (2) Concatenating original image helps to improve the precision of solving HSJP, because it provides a comparison for localising the initial positions of shuffled image patches.

As Figure \ref{cmpcontrast} demonstrates, we compare our performance of pretext (HSJP) and downstream (2D pose estimation) tasks under different settings: whether adding unshuffled image into input. Our experiments are also performed on HRNet W32. Unfortunately, although concatenating unshuffled image benefits the HSJP precision for pretext, it does not improve the performance of pose estimation downstream task: the mAP score is even worse than almost all weights pretrained without taking unshuffled images as inputs, and the performance is not better than training directly from scratch for 450 epochs. (In our scheme, the pretext HSJP task is trained for 240 epoch, the downstream task is trained for 210 epochs, so we try to train for 240+210=450 epochs from scratch for fair comparison.) Therefore, we do not concatenate the original image into input during HSJP pretraining, even if it seems to be reasonable.

\section{Conclusion}
In this paper, we propose self-supervised Heatmap-Style Jigsaw Puzzles (HSJP) problem as pretext task, for the pretraining of 2D human pose estimators. The target of HSJP is to learn the location of shuffled patches with a heatmap-based approach. We only use images of person instances in MS-COCO dataset during our HSJP pretraining, rather than introducing extra and much larger dataset, such as ImageNet. The weights learned by HSJP pretext task are utilised as backbones of 2D human pose estimator, which are then finetuned on MS-COCO human keypoints labels. Our downstream mAP scores are evaluated on two popular and strong 2D human pose estimators, HRNet and SimpleBaseline, and experimental results show that downstream pose estimators with our self-supervised pretraining obtain much better performance than those trained from scratch, and are comparable to those using ImageNet classification models as their initial backbones. However, we don't use any supervised data during our pretraining with HSJP pretext task, which saves the labour costs of labelling extra datasets (such as ImageNet) for pretraining.

For future works, our pretraining scheme with HSJP pretext task has potential to be used on other heatmap-based or pixel-labelling tasks, such as face keypoint detection, crowd counting, and semantic segmentation.

\newpage

{\small
\bibliographystyle{ieee_fullname}
\bibliography{egbib}
}

\begin{figure*}
\includegraphics{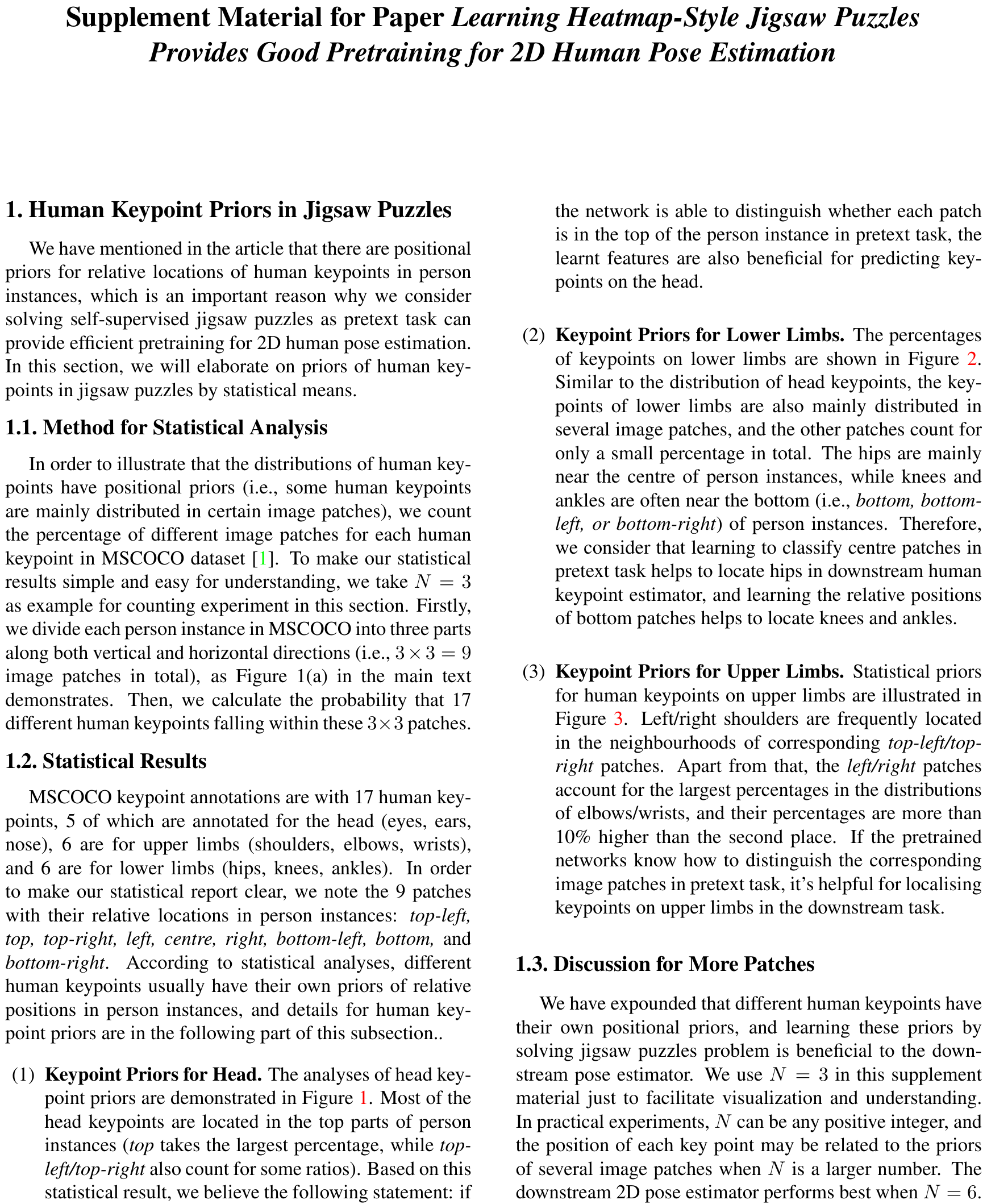}
\end{figure*}
\begin{figure*}
\includegraphics{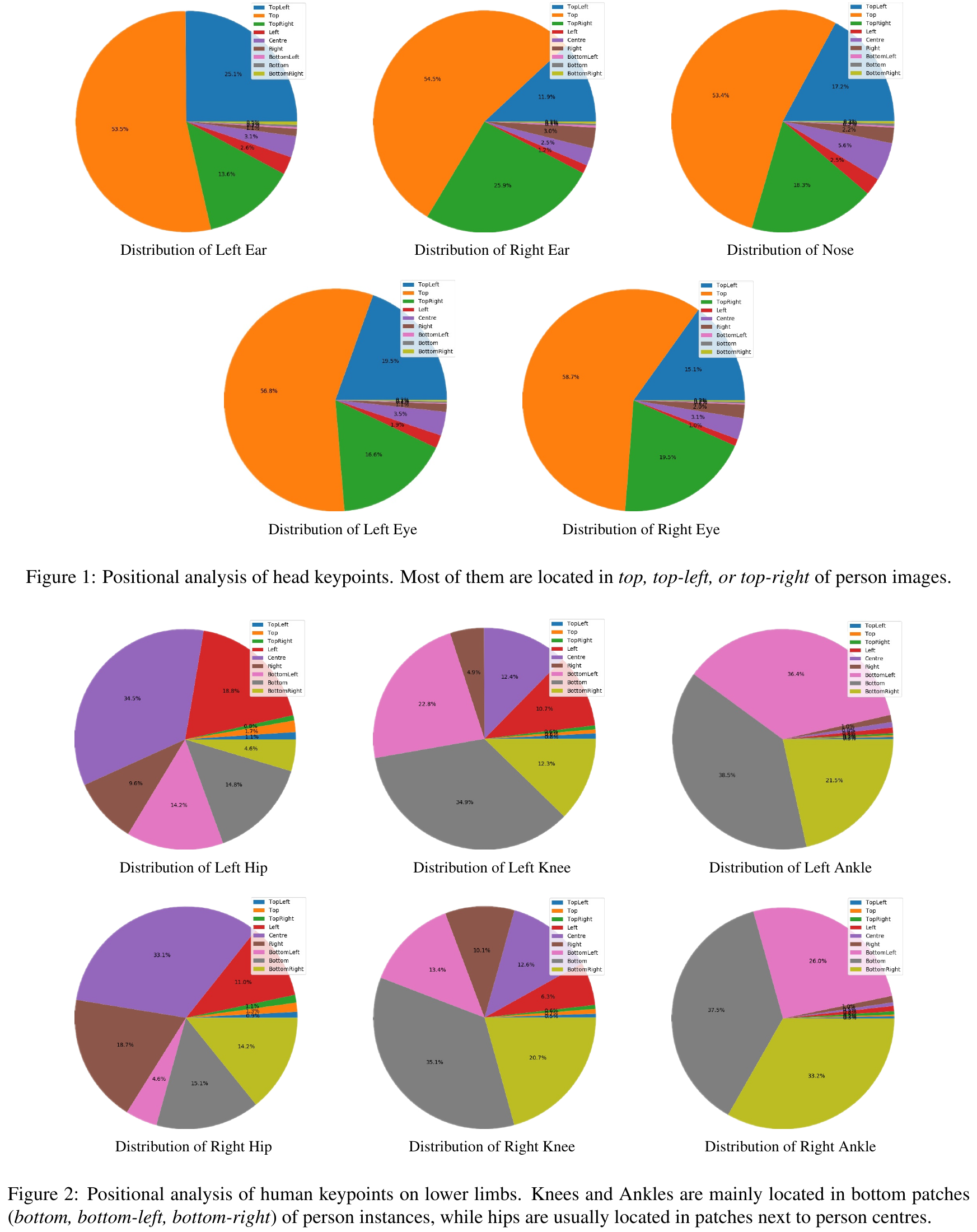}
\end{figure*}
\begin{figure*}
\includegraphics{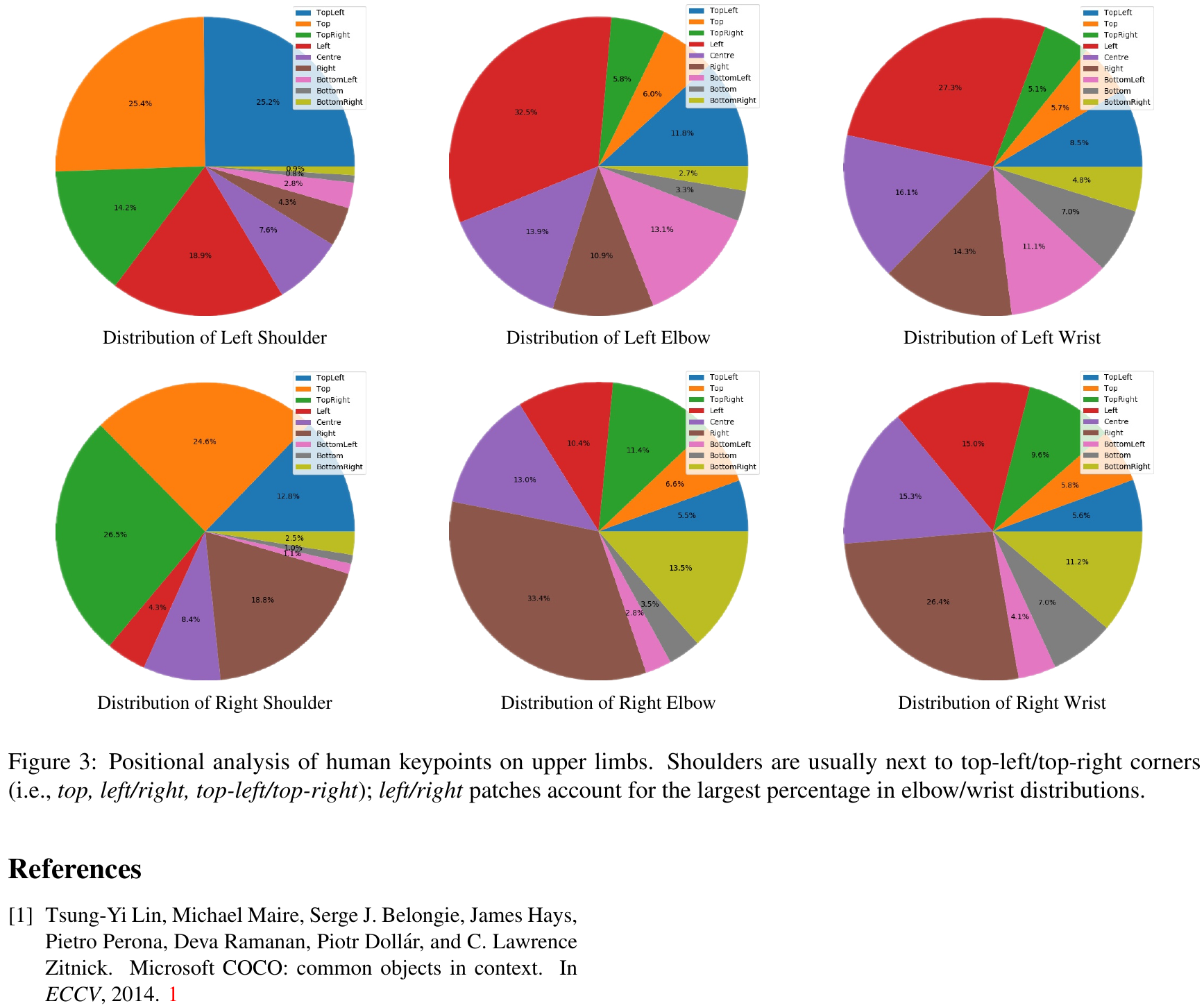}
\end{figure*}

\end{document}